\documentclass[conference]{IEEEtran}
\IEEEoverridecommandlockouts
\usepackage{cite}
\usepackage{algorithmic}
\usepackage{graphicx}
\usepackage{textcomp}

\usepackage{xspace,amsmath,amssymb,amsfonts,stmaryrd}
\usepackage{url}
\usepackage[defblank]{paralist}
\usepackage{tikz,subfigure}
\usetikzlibrary{arrows,shapes,shapes.multipart,snakes,automata,backgrounds,petri,positioning,shadows,matrix,decorations.pathmorphing,fit,positioning,calc,backgrounds}
\usepackage{todonotes}
\usepackage{xcolor}
\usepackage{multirow}
\usepackage{colortbl}
\usepackage{bm}
\usepackage{arydshln} 

\usepackage{fontawesome}

\usepackage[capitalise,nameinlink]{cleveref}

\newtheorem{example}{Example}
\newtheorem{definition}{Definition}
\definecolor{amber}{rgb}{1.0, 0.75, 0.0}

\crefname{section}{Sect.}{Sect.}
\Crefname{section}{Section}{Sections}
\crefname{example}{Ex.}{Ex.}
\Crefname{example}{Example}{Examples}
\crefname{figure}{Fig.}{Fig.}
\Crefname{figure}{Figure}{Figures}
\crefname{table}{Tab.}{Tab.}
\Crefname{table}{Table}{Tables}
\crefname{lstlisting}{List.}{List.}
\Crefname{lstlisting}{Listing}{Listings}
\crefname{definition}{Def.}{Defs.}
\Crefname{definition}{Definition}{Definitions}

\newcommand{\eg}{e.\,g.,\ }
\newcommand{\ie}{i.\,e.,\ }

\usepackage{csquotes} 

\newcommand{\labeltitle}[1]{\noindent\textbf{#1}.}

\newcommand{\qedwhite}{\hfill \ensuremath{\Box}}

\def\BibTeX{{\rm B\kern-.05em{\sc i\kern-.025em b}\kern-.08em
    T\kern-.1667em\lower.7ex\hbox{E}\kern-.125emX}}
       
\begin{document}

\newcommand{\dbnet}{DB-net\xspace}
\newcommand{\dbnets}{DB-nets\xspace}
\newcommand{\mmnet}{MM-net\xspace}
\newcommand{\mmnets}{MM-nets\xspace}

\newcommand{\md}{\textsl{MO}\xspace}
\newcommand{\mdb}{M_{db}}
\newcommand{\odb}{O_{db}}

\newcommand{\A}{\mathcal{A}}
\newcommand{\D}{\mathcal{D}}
\newcommand{\C}{\mathcal{C}}
\newcommand{\R}{\mathcal{R}}
\newcommand{\E}{\mathcal{E}}
\newcommand{\F}{\mathcal{F}}
\newcommand{\M}{\mathcal{M}}
\newcommand{\I}{\mathcal{I}}
\newcommand{\Y}{\mathcal{Y}}


\newcommand{\mathname}[1]{\ensuremath{\text{\textit{#1}}}}
\newcommand{\propername}[1]{\text{\textsf{\small #1}}\xspace}
\newcommand{\aop}[1]{\mathtt{#1}}


\newcommand{\set}[1]{\{#1\}}
\newcommand{\tup}[1]{\langle#1\rangle}
\newcommand{\mult}[1]{#1^\oplus}
\newcommand{\supp}[1]{\mathit{supp}(#1)}
\newcommand{\codom}[1]{\mathit{coDom}(#1)}
\newcommand{\true}{\mathsf{true}}
\newcommand{\false}{\mathsf{false}}

\newcommand{\qedboxfull}{\vrule height 5pt width 5pt depth 0pt}
\newcommand{\qedfull}{\hfill{\qedboxfull}}
\newcommand{\funsym}[1]{\mathtt{#1}}
\newcommand{\setsym}[1]{\mathit{#1}}
\newcommand{\restr}[2]{{
  \left.\kern-\nulldelimiterspace 
  #1 
  \vphantom{\big|} 
  \right|_{#2} 
  }}
\newcommand{\powerset}{\raisebox{.15\baselineskip}{\Large\ensuremath{\wp}}}
\newcommand{\naturals}{\mathbb{N}}

\newcommand{\typename}[1]{\mathbf{#1}}
\newcommand{\types}{\mathfrak{D}}
\newcommand{\type}{\mathcal{D}}
\newcommand{\dom}{\Delta}
\newcommand{\sigp}{\Gamma}
\newcommand{\sigf}{\Phi}
\newcommand{\interp}{\Sigma}

\newcommand{\ftype}{\funsym{type}}
\newcommand{\typesmd}{\mathfrak{D}_\md}
\newcommand{\vartype}{\funsym{type}}
\newcommand{\terms}{\mathcal{T}}
\newcommand{\cname}[1]{\mathsf{#1}}
\newcommand{\mmdb}[1]{\mathsf{mmdb}\mathsf{:}\mathsf{#1}}
\newcommand{\LITs}{\typename{L}}
\newcommand{\IRIs}{\typename{I}}


\newcommand{\source}{\funsym{src}}
\newcommand{\address}{\funsym{addr}}
\newcommand{\cast}[2]{#1{::}#2}
\newcommand{\mmlayer}{\mathcal{M}} 
\newcommand{\literals}{\dom_\typename{L}}
\newcommand{\iris}{\dom_\typename{I}}
\newcommand{\oid}{\typename{oid}}
\newcommand{\fname}[1]{\funsym{#1}}

\newcommand{\rdf}{\mathit{RDF}}
\newcommand{\constin}[1]{\setsym{Const}(#1)}
\newcommand{\varset}[1]{\mathcal{V}_{#1}}
\newcommand{\varsin}[1]{\setsym{Vars}(#1)}
\newcommand{\guards}{\mathcal{G}}
\newcommand{\queries}{\mathcal{Q}}
\newcommand{\bgp}{\textsc{BGP}}
\newcommand{\filter}{\textsc{Filter}}
\newcommand{\bind}{\textsc{Bind}}
\newcommand{\union}{\textsc{Union}}
\newcommand{\join}{\textsc{Join}}
\newcommand{\opt}{\textsc{Opt}}
\newcommand{\ans}{\ensuremath{\mathit{ans}}\xspace}
\newcommand{\evalp}[1]{\ensuremath{ \llbracket #1  \rrbracket}}
\newcommand{\map}{\theta}
\newcommand{\SELECT}{\bm{\mathsf{SELECT}}\xspace}
\newcommand{\FROM}{\bm{\mathsf{FROM}}\xspace}
\newcommand{\WHERE}{\bm{\mathsf{WHERE}}\xspace}
\newcommand{\FILTER}{\bm{\mathsf{FILTER}}\xspace}

\newcommand{\ASK}{\bm{\mathsf{ASK}}\xspace}
\newcommand{\CONSTRUCT}{\bm{\mathsf{CONSTRUCT}}\xspace}
\newcommand{\DESCRIBE}{\bm{\mathsf{DESCRIBE}}\xspace}

\newcommand{\pattern}{\mathcal{P}}

\newcommand{\actions}{\mathcal{A}}
\newcommand{\apar}[1]{{#1}{\cdot}\aop{params}}
\newcommand{\aadd}[1]{{#1}{\cdot}\aop{add}}
\newcommand{\adel}[1]{{#1}{\cdot}\aop{del}}
\newcommand{\actname}[1]{\textsc{#1}}
\newcommand{\doact}[2]{\funsym{apply}(#1,#2)}
\newcommand{\aset}{\mathit{F}^\aop{A}}
\newcommand{\dset}{\mathit{F}^\aop{D}}
\newcommand{\uset}{\mathit{F}^\aop{U}}
\newcommand{\mmtemplate}{mm-template}
\newcommand{\mmtemplates}{mm-templates}
\newcommand{\motemplate}{mo-template}
\newcommand{\motemplates}{mm-templates}
\newcommand{\subst}{\sigma}
\newcommand{\callf}[1]{ \triangleright #1}

\def\dbicon#1#2#3{
    \node at #1 [cylinder, shape border rotate=90, draw=white,fill=black,minimum height=#2,minimum width=#3,yshift=-.1cm] {};
}

\def\viewplace#1#2#3{
\begin{scope}[shift={#2}]
    \node (#1) at (0,0) [place,draw,label={#3}] {};
    \node at (0,0) [cylinder, shape border rotate=90, draw=white,fill=black,minimum height=.3cm,minimum width=.4cm,yshift=-.1cm] {};
\end{scope}
}

\newcommand{\nuvarset}{\Upsilon_{\types}}
\newcommand{\vars}{\mathcal{X}_{\types}}
\newcommand{\places}{P}
\newcommand{\cplaces}{P_c}
\newcommand{\vplaces}{P_v}
\newcommand{\transitions}{T}
\newcommand{\coloring}{\funsym{color}}
\newcommand{\quass}{\funsym{query}}
\newcommand{\guass}{\funsym{guard}}
\newcommand{\aass}{\funsym{act}}
\newcommand{\invars}[1]{\setsym{InVars}(#1)}
\newcommand{\outvars}[1]{\setsym{OutVars}(#1)}
\newcommand{\freshvars}[1]{\setsym{FreshVars}(#1)}
\newcommand{\inflow}{F_{in}}
\newcommand{\outflow}{F_{out}}
\newcommand{\net}{\mathcal{N}}
\newcommand{\tuples}[1]{\Omega_{#1}}
\newcommand{\snapshot}{\mathfrak{s}}
\newcommand{\enabled}[2]{#1[#2\rangle}
\newcommand{\firet}[3]{\enabled{#1}{#2}#3}
\newcommand{\val}{\setsym{Val}}
\newcommand{\snapshots}{\mathfrak{S}}
\newcommand{\tsys}[1]{\Lambda_{#1}}

 \definecolor{dbcolor}{HTML}{F39C12}
\definecolor{datacolor}{HTML}{FAD7A0}
\definecolor{activedatacolor}{HTML}{85C1E9}
\definecolor{viewcolor}{HTML}{5499C7}
\definecolor{actioncolor}{HTML}{2471A3}
\definecolor{netcolor}{HTML}{D576AE}
\definecolor{mgreen}{rgb}{0.128,0.428,0}
\definecolor{burntorange}{rgb}{0.8, 0.33, 0.0}
\definecolor{camouflagegreen}{rgb}{0.47, 0.53, 0.42}
\definecolor{copperrose}{rgb}{0.6, 0.4, 0.4}

\newcommand{\tname}[1]{{\fontfamily{phv}\selectfont {#1}}}
\newcommand{\pname}[1]{\textit{#1}\xspace}
\newcommand{\gbody}[1]{\color{mgreen}\funsym{#1}}
\newcommand{\gname}[1]{\color{mgreen}[\funsym{#1}]}

\tikzstyle{place}=[circle,thick,draw=black,fill=white,minimum size=7mm,font=\fontsize{9}{144}\selectfont]
\tikzstyle{transition}=[rectangle,thick,draw=black,fill=gray!20,minimum size=7mm]
\tikzstyle{enabledtransition}=[rectangle,very thick,draw=green!75,fill=green!20,minimum size=7mm]
 \tikzstyle{container}=[rectangle,rounded corners,very thick,draw=black!75,fill=black!20,minimum height=7mm,minimum width=14mm]
\tikzstyle{rplace}=[circle,ultra thick,draw=violet!75,fill=violet!20,minimum size=7mm]

\tikzstyle{erbox}=[draw, fill=gray!20, minimum width=7em, text width=6.0em, text centered,
  minimum height=3em,rounded corners,drop shadow]

\def\dbicon#1#2#3{
    \node at #1 [cylinder, shape border rotate=90, draw=white,fill=black,minimum height=#2,minimum width=#3,yshift=-.1cm] {};
}

\def\viewplace#1#2#3{
\begin{scope}[shift={#2}]
    \node (#1) at (0,0) [place,draw,label={#3}] {};
    \node at (0,0) [cylinder, shape border rotate=90, draw=white,fill=black,minimum height=.3cm,minimum width=.4cm,yshift=-.1cm] {};
\end{scope}
}

\def\place#1#2#3#4#5#6{
\begin{scope}[shift={#2}]
    \node (#1) at (0,0) [place,draw,label={#3:#4},label={#5:\{#6\}}] {};
\end{scope}
}

\def\vplace#1#2#3#4#5#6{
\begin{scope}[shift={#2}]
    \node (#1) at (0,0) [place,draw,label={#3:#4},label={#5:\{#6\}}] {};
    \node at (0,0) [cylinder, shape border rotate=90, draw=white,fill=black,minimum height=.3cm,minimum width=.4cm,yshift=-.1cm] {};
\end{scope}
}

\tikzstyle{placelem}=[draw,
    cloud,
    cloud puffs = 15,
    minimum width=.2cm,
    minimum height=.2cm,
    fill=white,
    thick]

\tikzstyle{netelem}=[draw,
    cloud,
    cloud puffs = 20,
    minimum width=1.8cm,
    minimum height=1.8cm,
    fill=blue!10,
    ultra thick]

\tikzstyle{relationselem}=[placelem,fill=pink!20]
\tikzstyle{noopelem}=[placelem,fill=orange!20]
\tikzstyle{enteredplace}=[place,fill=yellow!20]
\tikzstyle{boundplace}=[place,fill=yellow!30]
\tikzstyle{guardokplace}=[place,fill=yellow!40]
\tikzstyle{updatedplace}=[place,fill=yellow!50]
\tikzstyle{violplace}=[place,fill=red!10]
\tikzstyle{constrokplace}=[place,fill=green!10]
\tikzstyle{docommitplace}=[place,fill=green!30]
\tikzstyle{dorollbackplace}=[place,fill=red!30]
\tikzstyle{arc}=[-stealth',thick]
\tikzstyle{readarc}=[-,thick]


\tikzset{
state/.style={
       rectangle split,
       rectangle split parts=2,
       rectangle split part fill={red!30,blue!20},
       rounded corners,
       draw=black,  very thick,
       minimum height=2em,
       minimum width=2cm,
       inner sep=2pt,
       text centered,
       }
}

\title{Formalizing Integration Patterns with Multimedia Data}

\author{\IEEEauthorblockN{Marco Montali, Andrey Rivkin}
    \IEEEauthorblockA{Free University of Bozen-Bolzano\\
        \{lastname\}@inf.unibz.it} \and
    \IEEEauthorblockN{Daniel Ritter}
    \IEEEauthorblockA{SAP SE\\
    	daniel.ritter@sap.com}
    }
\maketitle

\begin{abstract}
The previous works on formalizing enterprise application integration (EAI) scenarios showed an emerging need for setting up formal foundations for integration patterns, the EAI building blocks, in order to facilitate the model-driven development and ensure its correctness. So far, the formalization requirements were focusing on more ``conventional'' integration scenarios, in which control-flow, transactional persistent data and time aspects were considered. However, none of these works took into consideration another arising EAI trend that covers social and multimedia computing. 
In this work we propose a Petri net-based formalism that addresses requirements arising from the multimedia domain. We also demonstrate realizations of one of the most frequently used multimedia patterns and discuss which implications our formal proposal may bring into the area of the multimedia EAI development.

\end{abstract}


\section{Introduction}
\label{sec:intro}
Recent business and socio-technical trends start relying on smart applications with advanced analysis techniques, the IoT, business and social networks~\cite{Ritter201736,ZimmermannPHW16,DBLP:journals/dagstuhl-manifestos/AbiteboulABBCD018}. This entails the need to employ enterprise application integration (EAI) for processing unstructured multimedia and semantic data, with concrete applications like
smart logistics, disease detection in agriculture and health-care, social sentiment analysis. The latter has been recently studied in the context of multimedia EAI in \cite{Ritter201736,RitterR17}.
More generally, the need for handling multimodel and knowledge-enriched data (incl. text and multimedia data) was also identified in the related data management \cite{DBLP:journals/dagstuhl-manifestos/AbiteboulABBCD018} and event-based processing domains (considering audio, video and social events) \cite{DBLP:journals/ivc/TzelepisMMIKBSY16}.


While multimedia integration solutions become more relevant and complex, solid formal foundations are crucial in order to ensure the behavioral correctness
  of multimedia EAI solutions (cf. \cite{Ritter201736}).
Such formal foundations had been given by formalizing the execution semantics of integration patterns \cite{hohpe2004enterprise,Ritter201736} -- the building blocks of EAI solutions -- using colored Petri nets (CPNs)  \cite{DBLP:conf/caise/FahlandG13} and (timed) \dbnets \cite{DBLP:conf/edoc/0001RMRS18,RITTER2019101439}. 
Still, results of these works 
do not apply to multimedia data, whereas the recent survey \cite{Ritter201736} identifies a lack of a suitable formalism for multimedia integration patterns.
\begin{example}
	\label{ex:example}
	\begin{figure}[bt]
		\centering
		\includegraphics[width=.9\columnwidth]{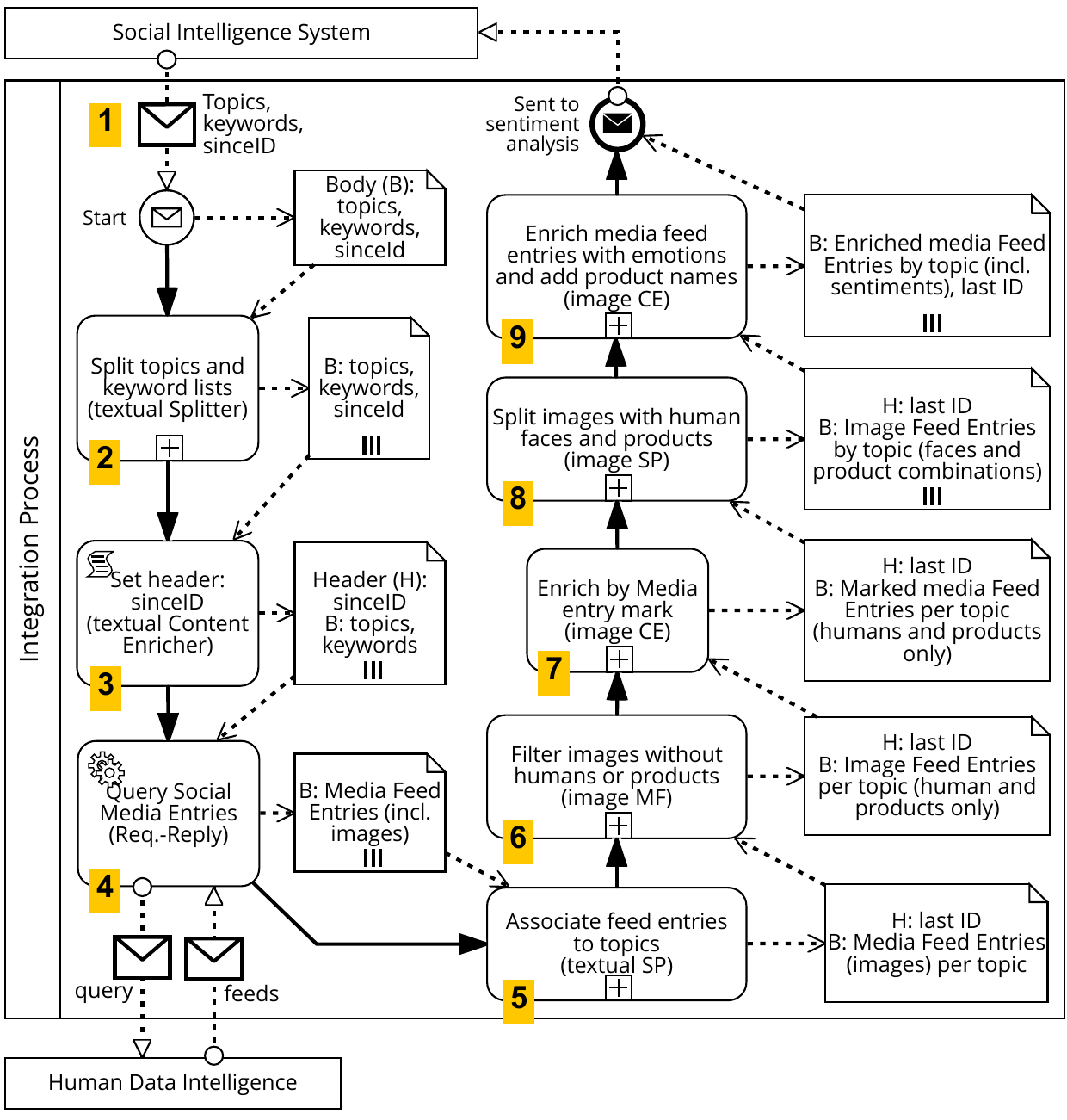}
		\caption{SAP Social Intelligence -- image sentiments (excerpt)}
		\label{fig:example}
		\vspace{-0.1cm}
	\end{figure}
	\cref{fig:example} shows an excerpt from a social media sentiment harvesting application (cf. \url{https://tinyurl.com/yautcagl}), in which images (and texts) are either collected from Human Data Intelligence providers or directly from social media sources like Facebook, Twitter, or Instagram. 
	For ease of reading, the diagram is enriched with numbers referring to concrete steps described in the text below.
	%
	To guide the search, a social intelligence system (\eg ERP, CRM) provides lists of \emph{topics} and \emph{keywords} of interest as well as time- or item-based metadata like a \emph{sinceID}, denoting the earliest feed of interest (see \textbf{{\colorbox{amber!70}1}}).
	Then, using textual Splitter and Content Enricher patterns (\textbf{{\colorbox{amber!70}2}} and \textbf{{\colorbox{amber!70}3}}), distinct queries are separated into multiple request messages with the sinceID as header (H).
	The resulting media feed entries contain images in the message body (B) that are processed by multiple subsequent steps (starting from \textbf{{\colorbox{amber!70}4}}).
	First, images without humans or products are filtered out using an image Message Filter (\textbf{{\colorbox{amber!70}6}}).
	Then an image Content Enricher marks the features (\textbf{{\colorbox{amber!70}7}}), along which the relevant parts in the images are split into separate messages by an image Splitter (\textbf{{\colorbox{amber!70}8}}).
	Finally, an image Enricher determines  the  emotional state of the human (towards the product) and adds the information to the image message, while  preserving  the  image (\textbf{{\colorbox{amber!70}9}}).
	The images with marked and determined sentiment as well as the association to the original topic are returned to the social intelligence system.
\hfill$\blacksquare$
\end{example}

In the absence of a formal representation of integration processes like the one in \cref{ex:example}, questions like \enquote{what does the process do?}, \enquote{is it functionally correct?} and \enquote{how can it be improved?} cannot be answered.
Consequently, reasoning about integration patterns with multimedia data (or even combined with textual data processing) is currently not possible, but desirable.
To answer these questions, this work combines the streams of previous research on EAI with multimedia data \cite{RitterR17} and formal representations of integration patterns on textual data \cite{DBLP:conf/caise/FahlandG13,DBLP:conf/edoc/0001RMRS18,RITTER2019101439} towards a novel formal representation of multimedia EAI solutions, which, apart from being defined using rigorous mathematical toolbox, should also allow to formally represent multimedia data in integration patterns and allow for further theoretical development along the line of formal analysis.
To this end, we build upon previous works on the EAI formalization using CPNs and \dbnets, and propose a new Petri net-based formalism called \emph{multimedia nets} (\mmnets for short). It can be essentially seen as marriage between CPNs and a multimedia storage whose conceptual representation is tuned to address various requirements specific to multimedia data management in the EAI context (\eg representation of multimedia messages, multimedia operations). 
%

In summary, the main \textbf{contributions} of this work along its outline are threefold.
\begin{inparaenum}[\it (1)] 
	\item First of all, we analyze multimedia data integration patterns regarding their requirements for defining a suitable formalism in \cref{sec:analysis}.
	\item Then, in \cref{sec:mmnets} we study formal syntax and semantics of the \mmnets and discuss certain design decisions behind the conceptual representation of multimedia data-related parts of the formalism.
	To the best of our knowledge, this is the first attempt to propose a formalism that would account for semantic knowledge and the way it is manipulated along a process execution.
	\item Finally, we give semantically correct realizations to one of the most frequently used multimedia integration patterns 
	in \cref{sec:realization-analysis}.
\end{inparaenum}
In \cref{sec:relatedwork} we discuss related work and conclude by briefly discussing further open research challenges in \cref{sec:conclusions}.

\section{Background and Requirements Analysis}
\label{sec:analysis}
In this section we briefly summarize multimedia integration patterns from which we derive requirements for a suitable formalization that we compare to the closest known related work on formalisms for textual integration patterns using CPNs \cite{DBLP:conf/caise/FahlandG13} and (timed) \dbnets \cite{DBLP:conf/edoc/0001RMRS18,RITTER2019101439}.

\subsection{Multimedia Integration Patterns}
\begin{figure}[bt]
	\centering
	\includegraphics[width=.75\columnwidth]{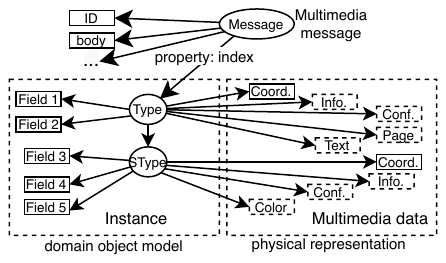}
	\vspace{-0.1cm}
	\caption{Conceptual Multimedia Message Model (from \cite{RitterR17})}
	\label{fig:coneptual}
\end{figure}

In previous work \cite{RitterR17}, we identified several integration patterns from the pattern catalogs \cite{hohpe2004enterprise,ritter2016exception,Ritter201736} that are especially relevant for multimedia data (cf. \cref{tab:imageOperations}).
\begin{table}[tb]
	\centering
	\scriptsize
	\caption{Integration pattern Multimedia Aspects, all information apart from \textbf{DB} taken from \cite{RitterR17} (logical -- \textbf{Log}, physical -- \textbf{Phy}, re-calculated -- recal., \textbf{DB} -- persist; \faThumbsOUp: yes, \faThumbsDown: no)}
	\vspace{-0.1cm}
	\label{tab:imageOperations}
	\begin{tabular}{lll|ll|l}
		\parbox[t]{1.0cm}{\textbf{Pattern Name}} & \parbox[t]{1.3cm}{\textbf{Multimedia Operation}} & \parbox[t]{2.3cm}{\textbf{Arguments}} & \textbf{Phy} & \textbf{Log} & \textbf{DB}\\
		\hline 
		\parbox[t]{1.0cm}{Channel Adapter} & \parbox[t]{1.3cm}{format conversion} & \parbox[t]{2.3cm}{format indicator} & write & create & \faThumbsDown \\
		\hdashline
		\parbox[t]{1.0cm}{Splitter} & \parbox[t]{1.3cm}{fixed grid, object-based} & \parbox[t]{2.3cm}{grid: horizontal, vertical cuts; object} & create & recal./write & \faThumbsDown\\ 
		\hdashline
		\parbox[t]{1.0cm}{Router, Filter} & \parbox[t]{1.3cm}{select object} & \parbox[t]{2.3cm}{object} & - & read & \faThumbsDown\\
		\hdashline
		\parbox[t]{1.0cm}{Aggregator} & \parbox[t]{1.3cm}{fixed grid, object-based} & \parbox[t]{2.3cm}{grid: rows, columns, heights, width} &  create & recal./write & \faThumbsOUp\\ 
		\hdashline
		\parbox[t]{1.0cm}{Translator, Content Filter} & \parbox[t]{1.3cm}{coloring} & \parbox[t]{2.3cm}{color (scheme)} & write & recal./write & \faThumbsDown \\ 
		\hdashline
		\parbox[t]{1.0cm}{Content Enricher} & \parbox[t]{1.3cm}{add shape, OCR text} & \parbox[t]{2.3cm}{object, shape+color, text} & write & recal./write & \faThumbsOUp\\ 
		\hdashline
		\parbox[t]{1.0cm}{Feature Detector} & \parbox[t]{1.3cm}{segmentation, matching} & \parbox[t]{2.3cm}{object classifier} & read & create & \faThumbsDown\\ 
		\hdashline
		\parbox[t]{1.0cm}{Image Resizer} & \parbox[t]{1.3cm}{scale image} & \parbox[t]{2.3cm}{size: height, width} & write & write & \faThumbsDown\\ 
		\hdashline
		\parbox[t]{1.0cm}{Idempotent Receiver} & \parbox[t]{1.3cm}{detector, similarity} & \parbox[t]{2.3cm}{object for comparison} & - & read & \faThumbsOUp\\
		\hdashline
		\parbox[t]{1.0cm}{Message Validator} & \parbox[t]{1.3cm}{detector} & \parbox[t]{2.3cm}{validation criteria} & - & read & \faThumbsDown\\
	\end{tabular}
\end{table}
In addition to the pattern name and the corresponding multimedia operation, the (semantic) configuration arguments relevant for modeling such patterns are added.
While, in general, the tasks of the multimedia patterns are similar to those working with textual data, they differ in terms of the message representation as well as performed operations and required storage (cf. \cite{RitterR17}).
A multimedia message consists of a body and an optional set of attachments, and both of them ``physically'' contain multimedia data (\eg image).
In addition to a set of key-value header entries denoting metadata concerning the data exchange (\eg HTTP headers), there is a set of properties that carries the semantics of the multimedia data (\eg human with emotion, product) in the message body (or attachments). 
In \cite{RitterR17} it is assumed that a multimedia message is \emph{transient} (\ie processed in a pipes-and-filter style), and that all operations are executed directly on media objects and their metadata contained in the message.
Moreover, for representing multimedia messages, \cite{RitterR17} adapts a concept from the multimedia database domain (\eg cf. \cite{DBLP:journals/tkde/ChangDPV07}), which separates the \emph{logical} and \emph{physical} representations to isolate the runtime from modeling, as it is abstractly reflected in the multimedia message model in \cref{fig:coneptual}.

The physical representation accounts for the actual multimedia data, and thus operations on the physical representation literally read (\ie \textsf{read}), create new (\ie \textsf{create}), or change existing multimedia data (\ie \textsf{write}) like cutting parts of or resizing an image.
When the physical representation is read and interpreted, semantic information is extracted (\eg a detected human emotion), together with additional information like coordinates of the detected object (\textsf{coord.}), its color and the confidence of the detection (cf. \textsf{Conf.}; \eg \textsf{type=``human''} with \textsf{Conf.=0.85}).
The detection is done by a Feature Detector pattern (cf. \cref{tab:imageOperations}) that has a set of ML-trained classifiers for each expected feature in multimedia data.
During the detection, the distinct features are identified using the classifiers and the corresponding logical representation gets created.
In contrast to \cite{DBLP:journals/tkde/ChangDPV07}, where a relational multimedia model is used, the semantic information is then represented logically as part of the domain object model with references to the physical representation.
This could be represented using, for example, the RDF standard~\cite{rdf} (which we rely on in a formalism presented in \cref{sec:mm-storage}). For example, \cref{fig:coneptual} denotes an abstract view of the domain object model by visually representing the semantic concepts as \textsf{Type} (\eg virtual human), with sub-types \textsf{SType} (\eg emotion). 
Note that such standards as XSD or WSDL, in which business domain objects are usually encoded (\eg business partner, customer, employee), are not sufficient (cf. \cite{RitterR17}) as they are used  for representing textual domain models.
During the modeling of a process, the user works close to the logical representation by implicitly using operations like read/query (\ie \textsf{read}), change (\ie \textsf{write}), newly create (\ie \textsf{create}), and adapt changes without new detection or creation (\ie \textsf{recal.}).
The aforementioned detector can be also used for cases when the physical representation has changed and the corresponding logical part is invalidated, thus requiring a re-detection (\textsf{write}).
However, for efficiency reasons, if the effect of the physical operation on the logical representation is known, then only a recalculation of the logical part can be used (\textsf{recal.}).
As such, the logical representation denotes a canonical data model based on the domain model or message schema of the multimedia messages as well as operations on them.
\cref{tab:imageOperations} shows the physical (\textbf{Phy}) and logical (\textbf{Log}) operations required by the different patterns as well as database information (\textbf{DB}) indicating whether a pattern requires persistent storage for its operation.
\begin{example} 
	The image Splitter uses an object-based splitting (cf. \cref{tab:imageOperations}), where the object is a human face.
	During the processing, new physical multimedia objects (\eg images of human faces in the original image) are created (\textbf{Phy}:create) and the logical representation is either created from the scratch (\textbf{Log}:create) or attempted to be recalculated (\textbf{Log}:recal.) according to the knowledge about the split.
	Notice that this does not require a persistent storage of multimedia messages.	\hfill$\blacksquare$
\end{example}

\subsection{Formalization Requirements}
The formalization requirements of multimedia integration patterns are derived from the patterns in \cref{tab:imageOperations}.
The base requirements (also found for integration patterns on textual data \cite{RITTER2019101439}) are necessary for representing the control flow that messages go through in the integration process (\ie \textbf{REQ-0 \enquote{control flow (pipes and filter)}}).
The next requirements concern the processing of multimedia data.
As discussed before, the data processing has two different aspects, dealing with the representation of multimedia messages, and physical and logical operations on the messages.

The representation of multimedia messages requires support for multimedia and semantic data, which is not provided by the CPN \cite{DBLP:conf/caise/FahlandG13} and (timed) \dbnet \cite{DBLP:conf/edoc/0001RMRS18,RITTER2019101439} approaches (\ie \textbf{REQ-1(a) \enquote{Multimedia message representation)}}).
Physical multimedia operations on the data require capabilities like marking an image with some geometrical shape for the Content Enricher (\ie \textbf{REQ-1(b) \enquote{Multimedia data operations}}), whereas the logical representation requires keeping the logical part ``up-to-date'' (\ie it does not describe features of a physical object that are not there).
This not only allows to efficiently process the data stored on the logical part, \eg by using SPARQL queries (cf. \cite{RitterR17}), but also to support the modeling, during which semantic operations on the multimedia data can be specified (\ie \textbf{REQ-1(c) \enquote{Semantic / metadata operations}}).
Another data processing aspect concerns the persistent storage for patterns like the Aggregator and Idempotent Receiver, which require the storage for their operations (\ie \textbf{REQ-1(d) \enquote{Persistently store multimedia data, semantic data / metadata}}).
In addition to the functional requirements, it is important to provide a suitable formal representation of such multimedia integration patterns so as to facilitate their correct representation and further development (\ie \textbf{REQ-2 \enquote{Formal rigorous semantics}}), as done in model-driven development~\cite{Broy09}. Notice that further requirements from \cite{RITTER2019101439} like time, transaction and exception handling are out-of-scope, since they are not directly related to multimedia data.
\cref{tab:requirements} summarizes the formalization requirements that we consider in this work by setting the coverage of two approaches based on colored Petri nets~\cite{DBLP:conf/caise/FahlandG13} and \dbnets~\cite{MR17,RITTER2019101439}, which, to date, are the only ones that have been used for formalising integration patterns.
While CPNs provide a solid foundation for control (cf. REQ-0) and a simple data flow representation, \dbnets extend the latter towards the support of persistent data 
with CRUD operations for working with external, transactional databases.
However, none of them supports multimedia data or semantic/metadata operations (REQ-1(b)--(c)).
CPNs do not support the modeling of persistent storage, whereas \dbnets do not allow for a conceptually correct representation of multimedia data as they cannot support two different storages (one for logical and the other for physical data) that would also need to be managed differently. 
As long as no multimedia or semantic data aspects are concerned, \dbnets can store that data (REQ-1(d)).
The well-defined semantics of CPNs and \dbnets allow to conduct various types of model-based analysis, ranging from model-based testing via simulation to complex verification using variants of temporal logics.

\begin{table}[tb]
	\caption{Formalization Requirements (covered: \faThumbsOUp, partially: (\faThumbsOUp), not: \faThumbsDown)}
	\label{tab:requirements}
		\small
	\begin{tabular}{llcc}
		\hline
		ID & Requirement & CPN & (timed) db-net \\ 
		\hline
		\hline
		REQ-0 & \parbox[t]{3.3cm}{Control flow (pipes and filter)} & \faThumbsOUp & \faThumbsOUp\\
		REQ-1 & (a) \parbox[t]{3.3cm}{Multimedia message representation} & \faThumbsDown & \faThumbsDown\\
		& (b) \parbox[t]{3.3cm}{Multimedia data operations} & \faThumbsDown & \faThumbsDown\\
		& (c) \parbox[t]{3.3cm}{Semantic / metadata operations} & \faThumbsDown & \faThumbsDown\\
		& (d) \parbox[t]{3.3cm}{Persistently store multimedia data, semantic data / metadata} & \faThumbsDown & (\faThumbsOUp)\\
		\hline
		REQ-2 & Formal rigorous semantics & \faThumbsOUp & \faThumbsOUp\\
		\hline
	\end{tabular}
\end{table}


\section{Multimedia nets}
\label{sec:mmnets}
\definecolor{Gray}{rgb}{0.66, 0.66, 0.66}

In this section we present the formalism of MM-nets that builds upon CPNs and takes inspiration from the multi-layered representation adopted within the \dbnet approach by subsequently defining multimedia data and semantic operations as well as multimedia storage for the data-related requirements (cf. REQ-1(a)--(d) from \cref{tab:requirements}).
Conceptually, \mmnets are structured as follows: 
\begin{inparaenum}[\it (i)] 
\item a \emph{multimedia storage} stores multimedia data together with their metadata; 
\item a \emph{control layer} employs a variant of CPNs to capture the control-flow dimension of the modeled process;
\item a \emph{data logic layer} embodies a communication interface between the multimedia and control layers. 
\end{inparaenum}
Using the data logic, the control layer can access the underlying multimedia storage (and tune its own behavior depending on the obtained answer) as well as update it with data carried by the tokens and additional data obtained from the external world. In what follows, we study every layer in detail and provide a formal definition of an \mmnet.
We also discuss how, in spite of certain conceptual and operational differences, \dbnets can be related to \mmnets. This observation provides insights on a possibility of adopting formal analysis techniques studied for \dbnets for the formalism of \mmnets. 


\subsection{Multimedia storage}
\label{sec:mm-storage}
A data type is $\type=\tup{\dom_\type,\sigp_\type,\sigf_\type,\interp_\type}$, where $\dom_\type$ is a value domain, $\sigp_\type$ and $\sigf_\type$ are finite sets of predicate and function symbols defined on top of elements of $\dom_\type$, $\interp_\type$ is the signature interpretation, \ie a function associating each predicate symbol $S$ (resp., function symbol $f$) of arity $n$, denoted as $S/n$ (resp., $f/n$), to an $n$-ary relation $\interp(S)\subseteq\dom_\type^n$ (resp., to an $n$-ary function $\interp(f):\dom_{\type_1}\times\ldots\dom_{\type_n}\rightarrow\dom_\type$, where each $\type_i$ is some type, possibly different from $\type$). For the sake of brevity, hereinafter we omit the signature interpretation in the data type definitions.

Examples of data types are:
\begin{inparaenum}[\it 1)]
\item $\typename{str}=\tup{\mathbb{S},\set{=_{s}},\emptyset}$ --  strings with the equality predicate;
\item $\typename{int}=\tup{\mathbb{Z},\set{=_{int}, <_{int}},\set{succ:\mathbb{Z}\rightarrow\mathbb{Z}}}$ -- integers with the usual comparison operators, as well as the successor function;
\item $\typename{jpg}=\tup{\mathbb{IMG},\emptyset,\set{\fname{sub}:\mathbb{IMG}\times\mathbb{IMG}\rightarrow\mathbb{IMG}}}$ -- images in JPG format with a binary image subtraction function.
\end{inparaenum} 
We use $\types$ to denote a type domain, that is, a finite set of data types, and write $\Box_\types=\bigcup_{\type\in\types}\Box_\type$, for $\Box\in\set{\dom,\sigp,\sigf}$.  Also, for ease of presentation, we single out a domain of multimedia (object) types $\typesmd$, s.t. $\typesmd\cap\types=\emptyset$, and fix a string-based type $\oid=\tup{\mathbb{S},\set{=_{oid}},\emptyset}\in\types$ for specifying proper addresses of objects.
Functions from $\sigf_{\typesmd}$ provide the basis for defining operations discussed in REQ-1(b)--(c). 
Finally, we shall use a function $\ftype$, defined on $\types\cup\typesmd$,  to return a data type of a variable or a value.

In this work we make a \emph{design decision} for modeling multimedia data in which one focuses on the object metadata and treats them as the ``first class citizen'' (\eg similar to \cite{DBLP:journals/tkde/ChangDPV07}), assuming that the actual (multimedia) objects are kept in some storage and can be accessed/manipulated only by references. In this case one should distinguish two different types of object manipulations. One type focuses on the way the objects are accessed and viewed/manipulated (e.g., accessing and resizing an image stored on some server), whereas the other considers auxiliary information about objects and thus allows for treating them in a more refined way. 
To account for the first type, we introduce \emph{object storage} (or database) $\odb$ that stores multimedia objects of various types. We shall not go into technical aspects of such database, but instead just assume that it provides functionality for adding and deleting multimedia data, and that every object can be accessed by using its proper addresses. 
W.l.o.g., we formally treat $\odb$ as a set of pairs $(a,mo)$, where $mo\in\odb$ is a multimedia object and $a$ is its address of type $\oid$,  
and assume that \emph{all the addresses are unique and two distinct objects can never be referenced by the same address}. For convenience, we  introduce two functions: $\address:\typesmd\rightarrow\dom_\oid$ that, given a multimedia object, returns its address, and $\source:\dom_\oid\rightarrow\typesmd$ that, given an address, returns an object that this address is pointing at.

The metadata of all the objects from $\odb$ together with their addresses are kept in \emph{metadata storage} $\mdb$ that is represented as an RDF graph -- a set of \emph{statements} $(s,p,o)$, with $s$ being a subject, $p$ being a predicate and $o$ being an object. 
Each statement triple is an atomic construct. Its subject describes an information resource,
  while its predicate represents a statement property referenced by an internationalized resource identifier (IRI) whose value is the statement \emph{object}. 
Note that, while $s$, $p$ and $o$ carry values of IRIs, $s$ and $o$ can also be RDF literals (for more details see~\cite{rdf}). 
Notably, IRIs as objects can be used to represent more complex, tree-structured values. 
The usage of IRIs in RDF statements is crucial as it allows for the unambiguous \emph{identification} of information resources. 
As opposed to \cite{rdf}, we do not use blank nodes and thus consider only ground RDF graphs. 
In what follows, we shall use $\literals$ to denote an infinite set of RDF literals and $\iris$ to denote an infinite set of IRIs, and we may collectively refer to both of them as \emph{RDF terms}. Here, $\typename{L}=\set{\literals,\sigp_\typename{L},\sigf_\typename{L}}$ and $\typename{I}=\set{\literals,\sigp_\typename{I},\sigf_\typename{I}}$ respectively denote datatypes of RDF literals and IRIs, where $\typename{L},\typename{I}\in\types$, and $\sigp$ and $\sigf$ are potentially nonempty sets of predicate and function symbols that we intentionally leave unspecified as their content depends on a concrete scenario (or, more specifically, on a used database management system).
Lastly, given the complexity of the data-type management in RDF, for ease of presentation we employ a  \emph{type casting function} ${::}$ that, given $x$ and a target type $t$, returns a value in $x$ that is cast to $t$. W.l.o.g., we assume the extension of this function on variables.

To query the multimedia storage we adopt SPARQL -- the standard W3C pattern-matching language for querying RDF graphs~\cite{sparql}. There are plenty of formal ways to define the syntax of SPARQL queries as well as the semantics of pattern evaluation. In this paper, instead, we only provide intuitions necessary for understanding how metadata are accessed and how SPARQL query answers can be manipulated in the context of the studied formalism of \mmnets. 
Let $\varset{\rdf}$ be an infinite set $\set{?x,?y,\ldots}$ of RDF variables, where for each $?x\in\varset{\rdf}$, $\ftype(?x)=\typename{I}\cup\typename{L}$.
The basic building block of SPARQL queries is a \emph{triple pattern} -- a tuple from $(\literals\cup\iris\cup\varset{\rdf})\times(\literals\cup\varset{\rdf})\times(\literals\cup\iris\cup\varset{\rdf})$. Finite sets of such tuples form \emph{basic graph patterns} (BGPs)~\cite{sparql}.
More complex graph patterns are inductively constructed from BGPs using various operations (e.g., OPT, JOIN, UNION) that are applicable to graph patterns and built-in conditions.
The semantics of graph patterns is defined in terms of partial functions  $\map: \varset{\rdf} \rightarrow \literals \cup \iris$ called \emph{mappings}.
Given a BGP $\pattern$, $\map(\pattern)$ denotes the BGP obtained by applying $\map$ to all variables in $\pattern$. We use a function $\varsin{\pattern}$ to denote the set of all variables in $\pattern$. Both $\map$ and $\setsym{Vars}$ can be easily extended to account for tuples of variables.

Given an RDF graph $\mdb$, the evaluation of a graph pattern $\pattern$ over $\mdb$ is specified as the set $\evalp{\pattern}_{\mdb}$ of mappings inductively defined using SPARQL operations and the BGP evaluation as the base case~\cite{PAG09,KRRXZ14}. Notably, for the pattern evaluation we use the simple entailment semantics, in which, in the base case, for every mapping $\map\in\evalp{\pattern}_{\mdb}$, it holds that $\map(\pattern)\subseteq\mdb$.
SPARQL queries use results of the pattern evaluation to form result sets and come in four different forms: $\SELECT$, $\ASK$, $\CONSTRUCT$ and $\DESCRIBE$~\cite{sparql}. However, in this work we are only interested in the first two. 
A $\SELECT$ query can be abstractly defined as $(\vec{w},\pattern)$, where $\pattern$ is a graph pattern and $\vec{w}=\tup{w_1,\ldots,w_k}$ is a vector of answer variables, such that $\set{w_1,\ldots,w_k}\subseteq\varsin{\pattern}$.
Such query is then evaluated over a graph $\mdb$ by applying mappings from 
$\map\in\evalp{\pattern}_{\mdb}$ to the variables in $\vec{w}$. We shall denote the resulting set (by default, SPARQL uses the bag-based semantics for the query evaluation~\cite{sparql}, but we opt for the set-based one) as $\ans(\mdb,\vec{w},\pattern)$.
An $\ASK$ query returns a boolean value indicating whether a pattern matches the given RDF graph and can be seen as a special case of a $\SELECT$ with an empty set of answer variables. In what follows, we use $\queries$ to define the set of all such SPARQL queries. 
\begin{example}
\label{ex:query1}
For brevity, assume an RDF vocabulary $\mathsf{mmdb}$ that standardizes all the metadata attributes as well as relations between them relevant to the scenario in \cref{ex:example}.
To extract a number of segments that contain human faces in every image in the multimedia storage together with that image identifier, 
we can use query $numSeg:=(\tup{?id,?c},\pattern_1)$, where $\pattern_1$ is defined as 
\begin{align*}
\SELECT~ ?id,?c~ \WHERE \{?id~ \mmdb{faceCount}~?c\}
\end{align*}
For accessing information about the segments with human faces, query $segs:=(\tup{?id,?seg},\pattern_2)$ is used. Pattern $\pattern_2$ it employs is defined as 
\begin{align*}
\SELECT~ ?id,?seg~ \WHERE \{?id~ \mmdb{faceSegment}~?seg\}
\end{align*}
Here, every segment object stores an alphanumeric string carrying two pairs of coordinates (\eg $\cname{"(100,120)..(205,205)"}$) to represent rectangle coordinates within which the segment is located.
\hfill$\blacksquare$
\end{example}

Since the multimedia storage essentially has no intensional part (i.e., there are no schemas either for object storage or metadata storage), we only define its extensional part. Formally, a \emph{$\typesmd$-typed multimedia storage instance} is a pair $(\mdb,\odb)$, where a $\mdb$ is a metadata storage instance and $\odb$ is a multimedia object storage instance. 
Here, each instance should be understood as a set of address-object pairs and object metadata observed at the given point in time.
Notice that this representation of metadata and multimedia data essentially fully meets REQ-1(a) and REQ-1(d).

\subsection{Data logic layer} 
\label{sec:data-logic}
Here we discuss how to manipulate the multimedia storage and show how to update metadata of objects stored in the multimedia storage (resp., object storage) by adding and deleting possibly multiple triples (resp., multimedia objects) at once. 
Such updates are realized by means of parametrized actions, each of which consists of a set of templates -- expressions that, once instantiated, assert which RDF triples (resp. multimedia objects) will be deleted from and added to the database.


We intend to provide two main types of operations for updating the multimedia storage. 
The first type works directly with the metadata storage and allows to add and delete a fixed number of triples. When using this type of updates, the modeler, however, should be aware that any changes of object metadata should faithfully reflect the actual state of the object itself. 
The second type allows to add, delete and/or update multimedia objects themselves. Sometimes such operations should be bundled with those of the first type so as to ensure the integrity of the object-metadata indivisibility principle.
We fix the infinite set of typed variables $\varset{\types}$, where for each $x\in\varset{\types}$, $\ftype(x)=\type$ and $\type\in\types$. Note that $\varset{\rdf}\subset\varset{\types}$. 

\begin{definition}\label{def:action}
A parameterized \emph{action} $\alpha$ is a triple $(\vec{p},\dset,\aset)$, where: 
\begin{compactenum}
\item $\vec{p}$ is a tuple of action formal parameters -- distinct variables from $\varset{\types}$;
\item $\dset=(\aop{mm^-},\aop{mo^-})$ and $\aset=(\aop{mm^+},\aop{mo^+})$ are two pairs such that:
\begin{compactitem}[$\bullet$]
\item $\aop{mm^-}$ and $\aop{mm^+}$ are finite sets of triples $(s,p,o)\in(\literals\cup\iris\cup Y)\times(\literals\cup Y)\times(\literals\cup\iris\cup Y)$ to be deleted from and added to the metadata storage, where $Y=\varsin{\vec{p}}\cap\varset{\rdf}$;
\item $\aop{mo^-}$ is a finite set of addresses $a$ of objects to be deleted from the object storage, where $a$ is either a constant from $\dom_\oid$ or a variable from $\varsin{\vec{p}}$ with $\ftype(a)=\oid$;
\item $\aop{mo^+}$ is a finite set of expressions $a\callf{f(x_1,\ldots,x_n)}$ generating objects to be added to the object storage, where $a$ is either a constant from $\dom_\oid$ or a variable of type $\oid$ from $\varsin{\vec{p}}$, $f\in\sigf_{\typesmd}$ with co-domain $\codom{f}\subset\dom_{\typesmd}$, and every $x_i$ is either a variable from $\varsin{\vec{p}}$ or a constant from $\types\cup\typesmd$. \qedwhite
\end{compactitem} 
\end{compactenum}
\end{definition}
Here, if an object with address $a$ is already present in the object storage, $a\callf{f(x_1,\ldots,x_n)}$ updates this object with the result of the function call. If, instead, there is no object with such an address, then the same expression adds a pair $(a,r)$ to the object storage, where $r$ is a result of the function call.
To access different components of $\alpha$, we make use of the following notation: $\apar \alpha=\vec{p}$, $\adel{\alpha}=\dset$, $\aadd{\alpha}=\aset$. 
Given a substitution $\subst:\varsin{\apar\alpha}\rightarrow \literals \cup \iris \cup \dom_\oid$ for $\apar{\alpha}$, an action instance $\alpha\subst$ is a ground action resulting by substituting parameters in $\alpha$ with corresponding values in $\subst$.\footnote{Note that for RDF triples, a notion of substitution coincides with the one of mapping with the only difference that the former is not partial.} 
An \emph{application} of $\alpha\subst$ to a multimedia storage instance $\I=(\mdb,\odb)$, denoted as $\doact{\alpha\subst}{\I}$, returns a new instance of the multimedia storage $\I'=(\mdb', \odb')$, such that: 
\begin{compactitem}[$\bullet$]
\item $\mdb'=(\mdb\setminus \aop{mm}_{\alpha\subst}^-)\cup \aop{mm}_{\alpha\subst}^+$, 
where 
$\aop{mm}_{\alpha\subst}^-=\hspace{-.2em}\bigcup\limits_{(s,p,o)\in \aop{mm^-}}\hspace{-.5em}\subst(s,p,o)$
and 
$\aop{mm}_{\alpha\subst}^+=\hspace{-.2em}\bigcup\limits_{(s,p,o)\in \aop{mm^+}}\hspace{-.5em}\subst(s,p,o)$;
\item $\odb'=(\odb\setminus (\aop{mo}_{\alpha\subst}^-\cup\aop{mo}_{\alpha\subst_1}^+))\cup \aop{mo}_{\alpha\subst_2}^+$, where, assuming for simplicity that $X=x_1,\ldots,x_n$,
$\aop{mo}_{\alpha\subst}^-=\hspace{-.2em}\bigcup\limits_{a\in \aop{mo^-}}(a,\source(\subst(a)))$,
$\aop{mo}_{\alpha\subst_1}^+=
\hspace{-.2em}\bigcup\limits_{\substack{a\callf{f(X)}\in \aop{mo^+},\\\exists o.(a,o)\in\odb}}\hspace{-.5em}(a,o)$
and 
$\aop{mo}_{\alpha\subst_2}^+=
\hspace{-.2em}\bigcup\limits_{a\callf{f(X)}\in \aop{mo^+}}\hspace{-.5em}(a,f(\subst(x_1),\ldots,\subst(x_n)))$.
\end{compactitem}
As in \dbnets~\cite{MR17}, in order to avoid situations in which the same fact is asserted to be added and deleted, we prioritize additions over deletions. The overall representation of actions and their semantics together with the ability to use type-specific functions allow to account for REQ-1(b)--(c).

\begin{example}
\label{ex:action1}
The Splitter (cf. \cref{ex:example}) employs an action called $\actname{getImage}$ that extracts sub-image $o'$ from image $o$ with address $a$, based on the information about segment $seg$ that identifies it, and generates relevant metadata about $o'$ (like name $n$, new address $a'$ and new  image identifier $id$) that are added to the metadata storage. This action uses five formal input parameters $\apar{\actname{getImage}}=\tup{a,seg,a',id,n}$ and performs the following updates. 
It only deletes the metadata from the original image that are related to the selected segment: $\adel{\actname{getImage}}=(\set{(\cast{id}{\LITs},\mmdb{faceSegment},\cast{seg}{\LITs})},\emptyset)$. 
Then, using $\aadd{\actname{getImage}}=(\aop{mm^+},\aop{mo^+})$, it adds all necessary metadata entries 
$\aop{mm^+}:=\set{(\cast{id}{\LITs},\mmdb{address},\cast{a'}{\LITs})), \\(\cast{id}{\LITs},\mmdb{format},\cast{\cname{.jpg}}{\LITs})), (\cast{id}{\LITs},\mmdb{name},\cast{n}{\LITs}) }$
and adds to the object storage an extracted image with 
$\aop{mo^+}=\set{ a\callf{\fname{extractIMG}(\source(a),seg)}}$.  Here,  $\fname{extractIMG}:\mathbb{IMG}\times\mathbb{D}\rightarrow \mathbb{IMG}$ takes as input an image and a rectangular selection ($\typename{rect}$ is a type defined on top of an alphanumeric set of rectangle coordinates $\mathbb{D}$ representing segments), and returns a subimage defined by the latter. 

To update image $(a,o)$ by cutting another image with address $a'$ from it, we define action $\actname{cutFromIMG}$, s.t., $\aadd{\actname{cutFromIMG}}=(\emptyset,\set{a\callf{\fname{sub}(\source(a),\source(a'))}})$.
Knowing that image $(a,o)$ is already in the storage, we use here the action ``updating'' semantics.
\qedwhite
\end{example}

Notice that Definition~\ref{def:action} allows to specify actions whose execution may be still inconsistent.
For example, one may delete an object without removing its metadata, which is intuitively not an expected type of behavior.


\subsection{Control layer}
Before defining the central notion of \mmnet, we fix some standard notions related to \emph{multisets}. For some set $A$, $\mult{A}:=\set{m:A\rightarrow \mathbb{N}}$ is the \emph{set of multisets} over $A$.
Given a multiset $S \in \mult{A}$, an element $a \in A$ and $n \in \mathbb{N}$, $S(a) \in \mathbb{N}$ denotes the number of times $a$ appears in $S$ and we write $a^n \in S$ if $S(a) = n$. Given $S_1,S_2 \in \mult{A}$, we define the following operations on multisets:
\begin{inparaenum}[\it (i)]
\item $S_1 \subseteq S_2$ (resp., $S_1 \subset S_2$) if $S_1(a) \leq S_2(a)$ (resp., $S_1(a) < S_2(a)$) for each $a \in A$;
\item $S_1 + S_2 = \set{a^n \mid a \in A \text{ and } n = S_1(a) + S_2(a)}$;
\item if $S_1 \subseteq S_2$, $S_2 - S_1 = \set{a^n \mid a \in A \text{ and } n = S_2(a) - S_1(a)}$;
\item given a number $k \in \mathbb{N}$, $k \cdot S_1 = \set{a^{kn} \mid a^n \in S_1}$;
\item $|m|=\sum_{a\in A}m(a)$.
\end{inparaenum}

A \mmnet net assigns to each place a color type, which in turn corresponds to a data type or to a cartesian product of multiple data types from $\types$. 
\emph{Inscriptions}, represented as tuples of variables from $\varset{\rdf}\cup\varset{\types}$, constants from $\dom_\types\cup\literals\cup\iris$ and \emph{terms} (constructed from functions from $\sigf_{\typesmd}\cup\sigf_{\types}$, variables and constants, and denoted as $\terms$), are used to reference contents of places. 
We denote by $\Omega_A$ the set of all possible inscriptions over a set $A$. 
Quite often, when manipulating various data objects, one would like to ensure provision of fresh data values (for example, a generation of globally fresh object identifiers). To this end, we adopt the well-known mechanism used in $\nu$-Petri nets~\cite{RVFE11} and introduce a countably infinite set $\nuvarset$ of $\types$-typed \emph{fresh variables}, where for every $\nu \in \nuvarset$, we have that $\dom_{\vartype(\nu)}$ is countably infinite (this provides an unlimited supply of fresh values). Hereinafter, we fix a countably infinite set of $\types$-typed variable $\vars = \varset{\types} \uplus \nuvarset$ as the disjoint union of ``normal" variables $\varset{\types}$ and fresh variables $\nuvarset$. 
Let us also introduce a \emph{guard} -- a formula defined as $\varphi::=S(x_1,\ldots,x_m)\,|\,\neg\varphi\,|\,\varphi\land\varphi\,|\,\top$, where $S\in\sigp_\type$ (for some $\type\in\types$) and $x_i$ is either a variable of type $\type$, or a constant from $\dom_{\type}$ . We use $\guards$ to denote a set of all possible guards.
Notice that guards are not defined on multimedia objects.

\begin{definition}
\label{def:control-mmnet}
A $\types$-typed \mmnet $\net$ is a tuple  
$(\types, \places,\transitions,\inflow,\outflow,\coloring,\quass,\guass,\aass)$, where:
\begin{compactenum}[$\bullet$]
\item $\places = \cplaces \cup \vplaces$ is a finite set of \emph{places} partitioned into control places $\cplaces$ and view places $\vplaces$ (decorated as \resizebox{.35cm}{!}{\tikz{\node [circle,draw,very thick,minimum width=.6cm, minimum height=.6cm] at (0,0) {}; \dbicon{(0,0)}{.4cm}{.3cm}}} and can connect to transitions only with read arcs);
\item $\transitions$ is a finite set of transitions, s.t. $P\cap T=\emptyset$;
\item $\coloring: \places \rightarrow \powerset(\types)$ is a place typing function;
\item $\quass: \vplaces \rightarrow \queries$ is a query assignment function, s.t., for every $p\in\vplaces$ with $\quass(p)=(\vec{w},\pattern)$, it holds that $\vartype(\vec{w})=\coloring(p)$;
\item $\inflow: \places \times \transitions \rightarrow \mult{\tuples{\varset{\types}}}$ is an input flow, s.t. $\vartype(\inflow(p,t))=\coloring(p)$ for every $(p,t)\in\places\times\transitions$;
\item $\guass:  \transitions \rightarrow \guards$ is a partial guard assignment function, s.t., for every $t\in\transitions$, 
$\varsin{\guass(t)}\subseteq\invars{t}$, where $\invars{t}=\cup_{p\in\places}\varsin{\inflow(p,t)}$;
\item $\outflow: \transitions \times \places \rightarrow \mult{\tuples{\vars \cup \dom_\types\cup\terms}}$ is an output flow, s.t. $\vartype(\outflow(t,p))=\coloring(p)$ for every $(t,p)\in\transitions\times\places$;
\item $\aass:\transitions\rightarrow\actions$ is a partial action assignment function, where $\actions$ is a finite set of actions.\qedwhite
\end{compactenum}
\end{definition}

Note that the given definition does not restrict the usage of objects in the guard formulas to only those from $\types$. In fact, one can even compare multimedia objects by using the $\source$ function. Inscriptions in the output flow can inject possibly fresh data via external variables that are not bound by any input inscription and that are taken from $\outvars{t}\setminus\invars{t}$, where $\outvars{t}=\cup_{p\in\places}\varsin{\outflow(t,p)}$ and every variable $x$ can be either from $\nuvarset$ or $\varset{\types}$.

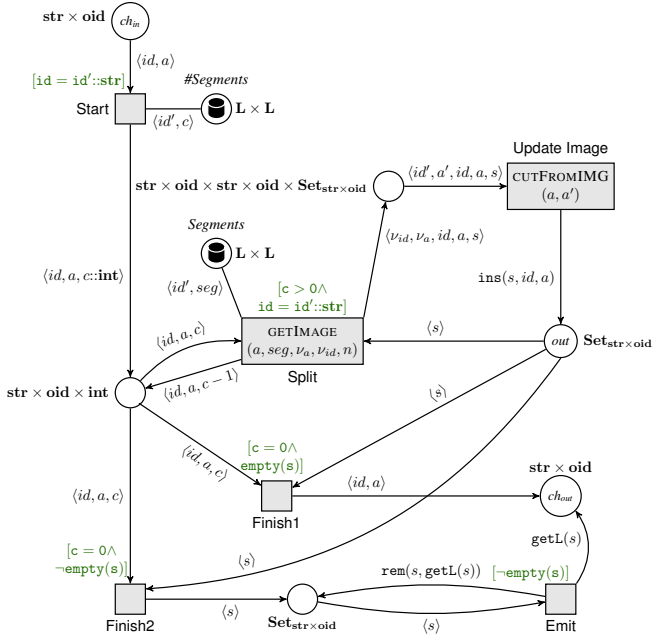
\begin{figure}[t!]
\centering
\resizebox{1\hsize}{!}{
\begin{tikzpicture}[->,>=stealth',auto,x=2cm,y=1.2cm,thick]

  \node[place,label={[yshift=1mm]left:$\typename{str}\times\typename{oid}$}] (in) at (1,5.2) {$\pname{ch}_\pname{in}$};

  \viewplace{objects}{(2,3.5)}{above,yshift=0cm:\pname{\#Segments}};
  \node[draw=none,right of = objects,xshift=-1mm,yshift=-0mm] (type) {$\typename{\LITs\times \LITs}$};

   \node[transition, label=left:\tname{Start}] (start) at (1,3.5) {};
  \node[draw=none,left of = start,xshift=-.2cm,yshift=6.5mm] (guard0) {$\gname{id=\cast{id'}{\typename{str}}}$};

  	
  	\node[place,label={[yshift=0mm]left:$\typename{str}\times\typename{oid}\times\typename{str}\times\typename{oid}\times\typename{Set}_{\typename{str}\times\typename{oid}}$}] (ready) at (4,2) {};
  	
	\node[transition, label=above:\tname{Update Image}] (update) at (6,2) {$\begin{array}{@{}c@{}} \actname{cutFromIMG}\\(a,a')\end{array}$};

	\viewplace{segments}{(2,.7)}{above,yshift=0cm:\pname{Segments}};
  	\node[draw=none,right of = segments,xshift=-1mm,yshift=0mm] (type) {$\typename{\LITs\times \LITs}$};
  
  	\node[place,label={[yshift=0mm]left:$\typename{str}\times\typename{oid}\times\typename{int}$}] (counter) at (1,-2) {};
  	
  	\node[transition, label=below:\tname{Split}] (extract) at (3,-1) {$\begin{array}{@{}c@{}} \actname{getImage}\\(a,seg,\nu_a,\nu_{id},n)\end{array}$};
  	\node[draw=none,above of = extract,xshift=0cm,yshift=0mm] (guard1) {$\begin{array}{@{}c@{}} \gbody{[c>0\land}\\ \gbody{id=\cast{id'}{\typename{str}}]}\end{array}$};
  	
  	  	\node[place,label={[yshift=0mm]right:$\typename{Set}_{\typename{str}\times\typename{oid}}$}] (set) at (6,-1) {\pname{out}};

  	\node[transition, label=below:\tname{Finish1}] (finish1) at (2.7,-4) {};
	\node[draw=none,above of = finish1,xshift=-1mm,yshift=-1mm] (guard2) {$\begin{array}{@{}c@{}}\gbody{[c=0\land}\\\gbody{\fname{empty}(s)]}\end{array}$};
	
	\node[place,label={[yshift=-1mm]above:${\typename{str}\times\typename{oid}}$}] (out) at (6,-4) {$\pname{ch}_\pname{out}$};
	  	
	\node[transition, label=below:\tname{Finish2}] (finish2) at (1,-6) {};
	\node[draw=none,above of = finish2,xshift=-9mm,yshift=-1mm] (guard3) {$\begin{array}{@{}c@{}}\gbody{[c=0\land}\\\gbody{\neg\fname{empty}(s)]}\end{array}$};
	  	
	\node[place,label={[yshift=1mm]below:$\typename{Set}_{\typename{str}\times\typename{oid}}$}] (p) at (3,-6) {};
	
	\node[transition, label=below:\tname{Emit}] (emit) at (6,-6) {};
	\node[draw=none,right of = emit,xshift=-17mm,yshift=6mm] (guard) {$\gname{\neg\fname{empty}(s)}$};

\path
(in) edge node[right,,pos=.4] {$\tup{id,a}$}  (start)
($(start.south)+(0,0)$) edge node[left,pos=.6]  {$\tup{id,a,\cast{c}{\typename{int}}}$}  (counter)

(counter) edge[bend left=25] node[above,xshift=-0mm,pos=.5,sloped] {$\tup{id,a,c}$} (extract.west)
(extract) edge node[below,xshift=1mm,pos=.5,sloped] {$\tup{id,a,c-1}$} (counter)
(extract.north east) edge node[right, pos=.7] {$\tup{\nu_{id},\nu_a,id,a,s}$}  (ready)
(set) edge node[above,pos=.6,sloped] {$\tup{s}$} (extract)
(set) edge node[above,pos=.4,sloped] {$\tup{s}$} (finish1)
(set.south) edge[bend left=25] node[above,pos=.8,sloped] {$\tup{s}$} ($(finish2.east)+(0,.2)$)
(ready) edge node[above, pos=.5] {$\tup{id',a',id,a,s}$}  (update)
(update) edge node[left,pos=.6] {$\fname{ins}(s,id,a)$} (set)
($(counter.south)+(.1,.1)$) edge node[below, pos=.6,sloped] {$\tup{id,a,c}$}  (finish1)
(counter) edge node[left, pos=.5] {$\tup{id,a,c}$}  (finish2)
(finish1) edge node[above, pos=.3] {$\tup{id,a}$}  (out)
(p) edge[bend right=10] node[below,pos=.5] {$\tup{s}$}  (emit)
(finish2) edge node[below,pos=.6] {$\tup{s}$} (p)
(emit) edge[bend right=10] node[above,pos=.5] {$\fname{rem}(s,\fname{getL}(s))$}  (p)
(emit.north east) edge[bend right=35] node[left, pos=.6] {$\fname{getL}(s)$}  (out.south east)
;

\path[-]
(objects) edge node[below,pos=0.5] {$\tup{id',c}$} (start)
(segments) edge node[left,pos=0.4] {$\tup{id',seg}$} (extract.north west)
;

\end{tikzpicture}
}
\caption{\mmnet control layer of a Splitter. Guards are depicted in green, types are shown in bold next to corresponding places.} \label{fig:mmnet_splitter}
\end{figure}
\begin{example}
\label{ex:net1}
A Splitter comprises a complex routing mechanism that, given a message, iteratively breaks it into smaller parts~\cite{hohpe2004enterprise}.
In our scenario we aim at splitting a single image into smaller images, shown in \cref{fig:mmnet_splitter}.
The splitting is performed according to a simplified criterion: only images of human faces will be extracted. All the  information needed for extracting such images is supposed to be already in the metadata storage. The latter is the key assumption guaranteeing that the Splitter can always identify needed elements and organize their processing. 

The net starts by extracting a number of segments containing human faces of all images in the multimedia storage (note that the information about the segments is supposed to be already in the metadata storage). This is done using view place \pname{{\#}Segments}, s.t., $\quass(\pname{{\#}Segments})=numSeg$, where $numSeg$ is defined in \cref{ex:query1}. By joining the input image identifier $id$ with the one in the view place, the net allows to get a number of segments $c$ for a concrete image. 
If no sub-images have been detected before, i.e., $c=0$, the net finishes its work by consequently firing transition \tname{Finish1}. Notice that \tname{Finish1} fires only if its guard has been satisfied, and that the inscription $\outflow(\text{\tname{Finish1}},\pname{out})$ contains a variable $s$ that is bound to a set of pairs (identifier, address) 
in place \pname{out}. Initially, this \pname{out} is supposed to contain an empty set.

If the image contains such segments ($c\neq 0$), the net enters a loop with $c$ as the counter and repeats the following steps. First, it fires \tname{Split} that executes action $\actname{getImage}$ assigned to it (cf. \cref{ex:action1}) that gets a sub-image based on the information of the concrete segment taken from view place \pname{Segments} (s.t., $\quass(\pname{Segments})=segs$, where $segs$ is defined in \cref{ex:query1}) and adds it to the multimedia storage together with the relevant metadata. 
$\actname{getImage}$ instantiates is formal parameters with variables that are bound to values coming from input places (like $a$ and $seg$) and that either simulate system input ($n$ is an unbounded variable simulating system input for an image name) or fresh data injection ($\nu_a$  creates a globally new image address, $\nu_{id}$ generates a fresh image ID). \tname{Split} also ``remembers'' the identifier and address of the extracted image by adding them to the set of pairs in place \pname{out}. 
Here, type $\typename{Set}_A$ (for $A:=A_1\times\ldots\times A_n$ and $\dom_A:=\dom_{A_1}\times\ldots\times \dom_{A_n}$) is defined over a set $\powerset(\dom_A)$, with predicate $\fname{empty}$ checking whether a set is empty or not, and three following functions:
\begin{inparaenum}[\it (i)]
\item $\fname{ins}$ adds an element to the set;
\item $\fname{getL}$ returns the last element from the set;
\item $\fname{rem}$ removes an element from the set.
\end{inparaenum}
The net then remembers the address of the extracted image and proceeds with firing the \tname{Update Image} transition that executes action $\actname{cutFromIMG}$ (cf. \cref{ex:action1}) that updates the image with address $a$ by removing from it the subimage with address $a'$. The loop repeats until the counter has reached $0$. Then the net fires \tname{Finish2} after which it consequently emits extracted pairs (using \tname{Emit}) into the output place $\pname{ch}_\pname{out}$.
\hfill$\blacksquare$ 
\end{example}


\subsection{Execution semantics} 
In the nutshell, the execution semantics of \mmnets is similar to the one of \dbnets~\cite{MR17}: it has to simultaneously capture the progression of both the multimedia storage and control layer. To this end, at each point in time, a state of a \mmnet is represented using a so-called \emph{snapshot}, that consists of multimedia storage instance $\I$ and \emph{marking} $m$. The latter is formally defined as function $m:\places\rightarrow \mult{\tuples{\types}}$, s.t. $m(p)\in\mult{\dom_{\coloring(p)}}$ and $m(v)=\ans(\mdb,\quass(v))$, for all $p\in\places$ and $v\in\vplaces$. Note that the second condition in the marking definition guarantees that the marking of a view place corresponds to the answers obtained by issuing its associated query over the underlying multimedia storage instance. In the following, by writing a \emph{\mmnet $\net$ in snapshot $\snapshot=(\I,m)$}, we mean a marked net, with marking $m$, over a multimedia storage instance $\I=(\mdb,\odb)$.

The firing of transition $t\in\transitions$ in a snapshot is defined w.r.t.
a so-called \emph{binding for $t$} defined as $\sigma:\varsin{t}\cup\outvars{t}\rightarrow \dom_\types$ that substitutes all variables in inscriptions on the arcs incident to $t$ and, possibly, formal parameters of an action signature assigned to $t$ with
values from $\dom_\types$. 
\begin{definition}
A transition $t\in\transitions$ is \emph{enabled in a snapshot $\snapshot=\tup{\I,m}$}, written as $\enabled{\snapshot}{t}$, if there exists a binding $\sigma$ satisfying the following:
\begin{inparaenum}[\it (i)]
\item $\sigma(\inflow(p,t))\subseteq m(p)$, for every $p\in\places$;
\item $\sigma(\guass(t))$ is true;
\item $\sigma(x)\not\in\val(\snapshot)$, for every $x\in\nuvarset \cap \outvars{t}$.\footnote{Here, $\val(\snapshot)$ denotes the set of all constants occurring both  in $m$ and $\I$.} \qedwhite

\end{inparaenum}
\end{definition}

Essentially, a transition is enabled with a binding $\sigma$ if the binding selects values carried by tokens from the input places (which match inscriptions on the corresponding input arcs), so that the data they carry make the guard attached to the transition true and, moreover, assigns globally fresh and pairwise distinct (both in $m$ and $I$) values to variables from $\nuvarset$. Then, when a transition is enabled, it may fire. 

\begin{definition}
Let $\net$ be a \mmnet in snapshot $\snapshot=(\I,m)$ (where $\I=(\mdb,\odb)$), with $t\in\transitions$ enabled in $\snapshot$ with some binding $\sigma$. Then, $t$ may fire producing new snapshot $\snapshot'=(\I',m')$, s.t. $m'(p)=m(p)-\sigma(\inflow(p,t))+\sigma(\outflow(t,p))$ and $\I'=\doact{\aass(t)\subst}{\I}$. We denote this as $\firet{\snapshot}{t}{\snapshot'}$ and assume that the definition is inductively extended to sequences $\tau\in\transitions^*$. \qedwhite
\end{definition}
For net $\net$ in initial snapshot $\snapshot_0$, we use $\snapshots(\net) = \set{\snapshot\mid \exists\tau\in T^* \text{, s.t.} | \firet{\snapshot_0}{t}{\snapshot}}$ to denote the set of all snapshots of $\net$ reachable from its initial snapshot $\snapshot_0$. 

\begin{figure}[t!]
\centering
\subfigure[]{
\resizebox{1\hsize}{!}{
\label{fig:mmnet_1}%
\begin{tikzpicture}[->,>=stealth',auto,x=1.8cm,y=1.2cm,thick]
  	
  	\viewplace{segments}{(0,2)}{above,yshift=0cm:\pname{Segments}};
  	\node[draw=none,left of = segments,xshift=1mm,yshift=0mm] (type) {$\typename{\LITs\times \LITs}$};

	\node[rectangle,rounded corners, draw,right of = segments,color=red!60,xshift=2.8cm] (token01) {$\begin{array}{@{}c@{}}\tup{\cname{84},\,\cname{(50,80)..(125,130)}}\\ \cdots\end{array}$};
	\node[circle, draw,fill=red!60,color=red!60,minimum size=5pt,inner sep=0pt] (token02) at (0,2) {} ;
	\draw[thick,draw=red!60,-] (token01) --(token02);
	
	\node[transition, label=left:\tname{Split}] (extract) at (0,0) {$\begin{array}{@{}c@{}} \actname{getImage}\\(a,seg,\nu_a,\nu_{id},n)\end{array}$};
  	\node[draw=none,above of = extract,xshift=-1.2cm,yshift=0mm] (guard1) {$\begin{array}{@{}c@{}} \gbody{[c>0\land}\\ \gbody{id=\cast{id'}{\typename{str}}]}\end{array}$};
  	
  	\node[place,label={[yshift=1mm]below:$\typename{str}\times\typename{oid}\times\typename{str}\times\typename{oid}\times\typename{Set}_{\typename{str}\times\typename{oid}}$}] (ready) at (2.5,0) {};
  	  	  	\node[draw = none, above of = ready,yshift=-4mm] (p2) { $\pname{ready}$};

	\node[transition, label=above:\tname{Update Image}] (update) at (4.85,0) {$\begin{array}{@{}c@{}} \actname{cutFromIMG}\\(a,a')\end{array}$};

  	\node[place,label={[yshift=0mm]right:$\typename{Set}_{\typename{str}\times\typename{oid}}$}] (set) at (2.5,-2.5) {};
  	  	\node[draw = none, left of = set,xshift=3mm] (out) { $\pname{out}$};

  	\node[rectangle,rounded corners, draw,below of = set,color=blue!60,yshift=-1mm] (token41) {$\tup{\set{~}}$};
	\node[circle, draw,fill=blue!60,color=blue!60,minimum size=5pt,inner sep=0pt] (token42) at (2.5,-2.5) {} ;
	\draw[thick,draw=blue!60,-] (token41) --(token42);
  	
  	\node[place,label={[yshift=0mm]right:$\typename{str}\times\typename{oid}\times\typename{int}$}] (counter) at (0,-2.5) {};
  	  	  	\node[draw = none, left of = counter,xshift=1.2mm] (p1) { $\pname{count}$};

	\node[rectangle,rounded corners, draw,below of = counter,color=blue!60,yshift=-1mm] (token11) {$\tup{\cname{84},\,\cname{images/id/84/party2.jpg},\,\cname{4}}$};
	\node[circle, draw,fill=blue!60,color=blue!60,minimum size=5pt,inner sep=0pt] (token12) at (0,-2.5) {} ;
	\draw[thick,draw=blue!60,-] (token11) --(token12);
\path
(counter) edge[bend left=25] node[left,xshift=-0mm,pos=.5] {$\tup{id,a,c}$} (extract)
(extract) edge[bend left=25] node[right,xshift=1mm,pos=.8] {$\tup{id,a,c-1}$} (counter)
(extract.east) edge node[above, pos=.5] {$\tup{\nu_{id},\nu_a,id,a,s}$}  (ready)
(ready) edge node[above, pos=.5] {$\tup{id',a',id,a,s}$}  (update)
(update) edge node[above,pos=.6,sloped] {$\fname{ins}(s,id',a')$} (set)
(set) edge node[above,pos=.4,sloped] {$\tup{s}$} (extract)
;

\path[-]
(segments) edge node[right,pos=0.4,xshift=-1mm] {$\tup{id',seg}$} (extract)
;
\end{tikzpicture}
}
}
\begin{tikzpicture}[->,>=stealth',auto,node distance=2.0cm,
  thick]
 \begin{scope}[xshift=2.5cm]
 \draw [->, ultra thick,snake=snake,segment amplitude=.4mm,segment length=2mm,line after snake=1mm] (1,.5) -- node[right,xshift=2mm] {$\begin{array}{@{}c@{}}\text{firing of } t \\\text{ with binding }\sigma\end{array}$} node[left,xshift=-2mm] {
\begin{tabular}{@{}r@{\ }c@{\ }l@{}}
$\sigma(id)$&$=$&$\cname{84}$\\
$\sigma(a)$&$=$&$\cname{images/id/84/party2.jpg}$\\
$\sigma(c)$&$=$&$\cname{4}$\\
$\sigma(seg)$&$=$&$\cname{(50,80)..(125,130)}$\\
$\sigma(\nu_{id})$&$=$&$\cname{84f1}$\\
$\sigma(\nu_a)$&$=$&$\cname{images/id/84/party2\_f1.jpg}$\\
$\sigma(n)$&$=$&$\cname{party2\_face1.jpg}$\\
\end{tabular}}  (1,-3.1);
 \end{scope}
\end{tikzpicture}
\subfigure[]{
\resizebox{1\hsize}{!}{
\label{fig:mmnet_2}%
\begin{tikzpicture}[->,>=stealth',auto,x=1.8cm,y=1.2cm,thick]
  	\node[place,label={[yshift=0mm]right:$\typename{str}\times\typename{oid}\times\typename{int}$}] (counter) at (0,-2.5) {};
  	 \node[draw = none, left of = counter,xshift=1.2mm] (p1) { $\pname{count}$};

  	  	\viewplace{segments}{(0,2)}{above,yshift=0cm:\pname{Segments}};
  	\node[draw=none,left of = segments,xshift=1mm,yshift=0mm] (type) {$\typename{\LITs\times \LITs}$};

  	\node[rectangle,rounded corners, draw,right of = segments,color=red!60,xshift=1.2cm] (token01) {$\begin{array}{@{}c@{}}\ \cdots~\end{array}$};
	\node[circle, draw,fill=red!60,color=red!60,minimum size=5pt,inner sep=0pt] (token02) at (0,2) {} ;
  		\draw[thick,draw=red!60,-] (token01) --(token02);

	\node[transition, label=left:\tname{Split}] (extract) at (0,0) {$\begin{array}{@{}c@{}} \actname{getImage}\\(a,seg,\nu_a,\nu_{id},n)\end{array}$};
  	\node[draw=none,above of = extract,xshift=-1.2cm,yshift=0mm] (guard1) {$\begin{array}{@{}c@{}} \gbody{[c>0\land}\\ \gbody{id=\cast{id'}{\typename{str}}]}\end{array}$};
  	
  	\node[place,label={[yshift=1mm]below:$\typename{str}\times\typename{oid}\times\typename{str}\times\typename{oid}\times\typename{Set}_{\typename{str}\times\typename{oid}}$}] (ready) at (2.5,0) {};
  	  	  	  	\node[draw = none, above of = ready,yshift=-4mm] (p2) { $\pname{ready}$};

	\node[transition, label=above:\tname{Update Image}] (update) at (4.85,0) {$\begin{array}{@{}c@{}} \actname{cutFromIMG}\\(a,a')\end{array}$};

  	\node[place,label={[yshift=0mm]right:$\typename{Set}_{\typename{str}\times\typename{oid}}$}] (set) at (2.5,-2.5) {};
  	  	\node[draw = none, left of = set,xshift=3mm] (out) { $\pname{out}$};

\node[rectangle,rounded corners, draw,below of = counter,color=blue!60,yshift=-1mm] (token11) {$\tup{\cname{84},\,\cname{images/id/84/party2.jpg},\,\cname{3}}$};
	\node[circle, draw,fill=blue!60,color=blue!60,minimum size=5pt,inner sep=0pt] (token12) at (0,-2.5) {} ;
		\draw[thick,draw=blue!60,-] (token11) --(token12);

		\node[rectangle,rounded corners, draw,above of = ready,color=blue!60,yshift=3.5mm] (token21) {$\begin{array}{@{}c@{}}\langle\cname{84f1},\,\cname{images/id/84/party2\_face1.jpg},\\\cname{84},\,\cname{images/id/84/party2.jpg},\,\cname{3},\,\cname{\set{~}}\rangle\end{array}$};
	\node[circle, draw,fill=blue!60,color=blue!60,minimum size=5pt,inner sep=0pt] (token22) at (2.5,0) {} ;
		\draw[thick,draw=blue!60,-] (token21) --(token22);

\path

(counter) edge[bend left=25] node[left,xshift=-0mm,pos=.5] {$\tup{id,a,c}$} (extract)
(extract) edge[bend left=25] node[right,xshift=1mm,pos=.8] {$\tup{id,a,c-1}$} (counter)
(extract.east) edge node[above, pos=.5] {$\tup{\nu_{id},\nu_a,id,a,s}$}  (ready)
(ready) edge node[above, pos=.5] {$\tup{id',a',id,a,s}$}  (update)
(update) edge node[above,pos=.6,sloped] {$\fname{ins}(s,id',a')$} (set)
(set) edge node[above,pos=.4,sloped] {$\tup{s}$} (extract)
;

\path[-]
(segments) edge node[right,pos=0.4,xshift=-1mm] {$\tup{id',seg}$} (extract)
;
\end{tikzpicture}
}
}
\caption{A marked \mmnet \cref{fig:mmnet_1} and the marked net \cref{fig:mmnet_2} resulting from the firing of a transition with a given binding} \label{fig:mmnet-firing}
\end{figure}
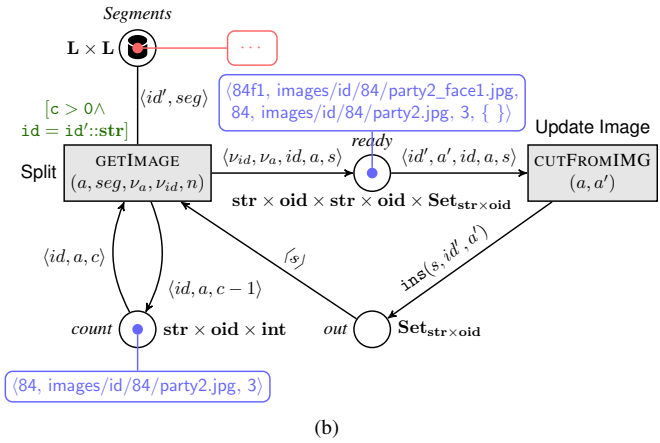
\begin{example}
\label{ex:net-firing}
\cref{fig:mmnet-firing} illustrates the effect of firing transition \tname{Split} with a given binding of a marked \mmnet. Now we briefly describe parts of the multimedia storage instances relevant to the nets in \cref{fig:mmnet-firing}. The first net in \cref{fig:mmnet_1} makes use of a multimedia storage instance (that, for our convenience, we denote as $\I_1$), in which metadata storage $\mdb^1$ primarily consists of RDF-triples describing the image with identifier $\cname{84}$: 

\begin{center}
\vspace{2mm}
\begin{tabular}{ l} 
 \hline
\rowcolor{Gray}
\multicolumn{1}{c}{$\mdb^1$} \\ 
 \hline\small
$\tup{\cname{84},\,\mmdb{format},\,\cname{.jpg}}$ \\ 
 $\tup{\cname{84},\,\mmdb{name},\,\cname{party2}}$   \\ 
 $\tup{\cname{84},\,\mmdb{address},\,\cname{images/id/84/party2.jpg}}$  \\
  $\tup{\cname{84},\,\mmdb{faceCount},\,\cname{4}}$ \\ 
 $\tup{\cname{84},\,\mmdb{faceSegment},\,\cname{(50,80)..(125,130)}}$\\
 $\tup{\cname{84},\,\mmdb{faceSegment},\,\cname{(204,80)..(270,130)}}$\\
\multicolumn{1}{c}{$\cdots$} \\
 \hline
\end{tabular}
\vspace{2mm}
\end{center}

The object storage of $\I_1$, denoted as $\odb^1$, contains an image with address $\cname{images/id/84/party2.jpg}$, in which all human faces have been already detected and suitably marked with red rectangles (notice that these rectangles should correspond to coordinates appearing RDF-triples with predicate $\mmdb{faceSegment}$):

\begin{center}
\vspace{2mm}
\begin{tabular}{cl} 
 \hline
\rowcolor{Gray}
\multicolumn{2}{c}{$\odb^1$} \\ 
 \hline\small
& \multirow{6}*{\begin{minipage}{.1\textwidth}\vspace{1mm}
      \includegraphics[height=25mm] {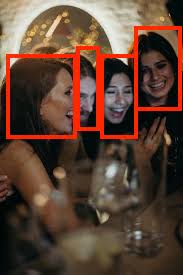} 
    \end{minipage}}\\ 
    &\\&\\ $\cname{images/id/84/party2.jpg}$&\\&\\&\\
    \multicolumn{2}{c}{$\cdots$} \\
 \hline
\end{tabular}
\vspace{2mm}
\end{center}

By firing \tname{Split}, action $\actname{getImage}$ assigned to it (as by its definition in \cref{ex:action1}) removes an RDF-triple that contains coordinates of the segment corresponding to the first face and inserts RDF-triples containing metadata of the newly extracted image (with identifier $\cname{84f1}$) of that face. The metadata storage component of the new multimedia storage instance, that, w.l.o.g., we name $\I_2$, looks as follows:  

\begin{center}
\vspace{2mm}
\begin{tabular}{ l} 
 \hline
\rowcolor{Gray}
	\multicolumn{1}{c}{$\mdb^2$} \\ 
 \hline\small
$\tup{\cname{84},\,\mmdb{format},\,\cname{.jpg}}$ \\ 
 $\tup{\cname{84},\,\mmdb{name},\,\cname{party2}}$   \\ 
 $\tup{\cname{84},\,\mmdb{address},\,\cname{images/id/84/party2.jpg}}$  \\
  $\tup{\cname{84},\,\mmdb{faceCount},\,\cname{4}}$ \\ 
 $\tup{\cname{84},\,\mmdb{faceSegment},\,\cname{(204,80)..(270,130)}}$\\
 $\tup{\cname{84f1},\,\mmdb{format},\,\cname{.jpg}}$ \\ 
  $\tup{\cname{84f1},\,\mmdb{name},\,\cname{party2-f1}}$   \\ 
   $\tup{\cname{84f1},\,\mmdb{address},\,\cname{images/id/84/party2\_f1.jpg}}$  \\
\multicolumn{1}{c}{$\cdots$} \\
 \hline
\end{tabular}
\vspace{2mm}
\end{center}
The multimedia object storage component of $\I_2$ will contain, above all, the newly extracted image together with its address: 

\begin{center}
\vspace{2mm}
\begin{tabular}{cl} 
 \hline
\rowcolor{Gray}
\multicolumn{2}{c}{$\odb^2$} \\ 
 \hline\small
& \multirow{6}*{\begin{minipage}{.1\textwidth}\vspace{1mm}
      \includegraphics[height=25mm] {images/party1} 
    \end{minipage}}\\ 
    &\\&\\ $\cname{images/id/84/party2.jpg}$&\\&\\&\\
    
& \multirow{6}*{\begin{minipage}{.1\textwidth}\vspace{1mm}
      \includegraphics[height=20mm] {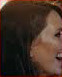} 
    \end{minipage}}\\ 
    &\\&\\ $\cname{images/id/84/party2f1.jpg}$&\\&\\&\\
    \multicolumn{2}{c}{$\cdots$} \\
 \hline
\end{tabular}
\vspace{2mm}
\end{center}

Notice that $\I_2$ makes part of the snapshot of the net in \cref{fig:mmnet_2}. $\blacksquare$ 

\end{example}

To address the last requirement from \cref{tab:requirements} (\ie REQ-2), we discuss the execution semantics of \mmnets. 
The execution semantics of a \mmnet is defined in terms of a possibly infinite-state labeled transition system (LTS) accounting for all possible executions of the control layer starting from an initial snapshot. States of this transition systems are \mmnet snapshots, whereas transitions model firings of \mmnet transitions under chosen bindings. Formally, given a \mmnet $\net$ in snapshot $\snapshot_0$, the execution semantics of $\net$ is given by the LTS $\tsys{\net}=(S,\snapshot_0,\rightarrow)$, where:
\begin{compactitem}[$\bullet$]
\item $S$ is a possibly infinite set of snapshots;
\item $\rightarrow\subseteq S\times \transitions \times S$ is a $\transitions$-labelled transition relation between pairs of snapshots;
\item $S$ and $\rightarrow$ are defined by simultaneous induction as the smallest sets satisfying the following conditions:
\begin{inparaenum}[\it(i)]
\item $\snapshot_0\in S$;
\item given $\snapshot\in S$, for every transition $t\in\transitions$, binding $\sigma$ and snapshot $\snapshot'$ over $\net$, if $\firet{\snapshot}{t}{\snapshot'}$, then $\snapshot'\in S$ and $\snapshot\overset{t}{\rightarrow} \snapshot'$.
\end{inparaenum}
\end{compactitem}
To make an LTS of some \mmnet visually more compact, one may omit the multimedia object storage component. This could also be enforced already in the above definition as any treatment of multimedia objects is done by using their addresses that, in turn, should be always present in the net's metadata storage.

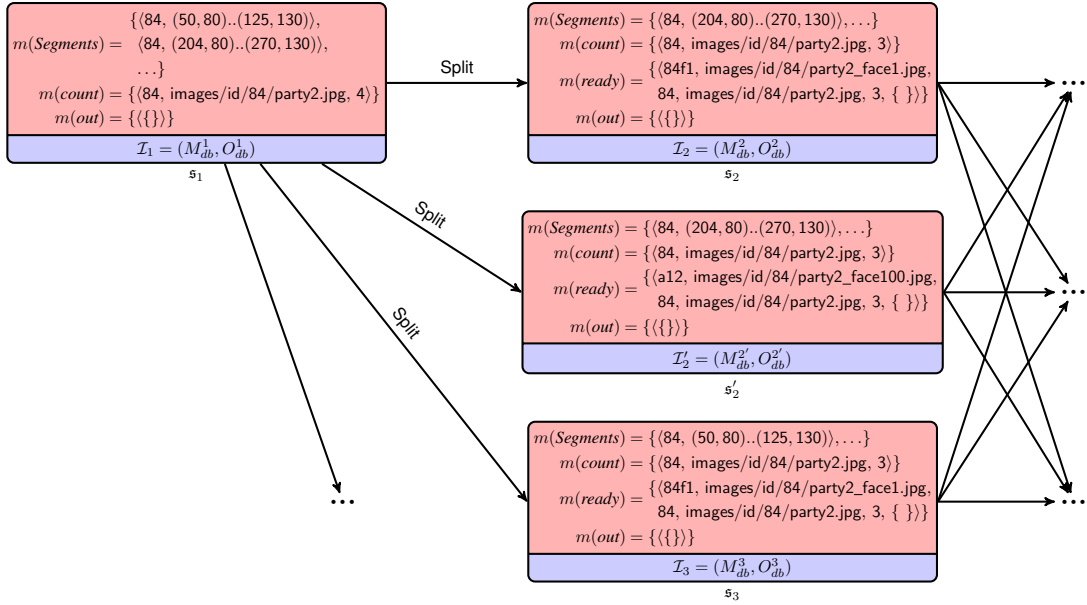
\begin{figure*}[t!]
\centering
\resizebox{.8\hsize}{!}{
\begin{tikzpicture}[->,>=stealth',auto,x=15cm,y=1.2cm,thick]
  	\arraycolsep=1.2pt\def\arraystretch{1.2}

  	\node[state, very thick,label=below:$\snapshot_1$] (s1) at (0,0) {%
		$\begin{array}{rl}
		m(\pname{Segments})&=\begin{array}{l}
													\{\tup{\cname{84},\,\cname{(50,80)..(125,130)}},\\
													\;\, \tup{\cname{84},\,\cname{(204,80)..(270,130)}},\\
													\;\, \ldots\}
											\end{array}\\
		m(\pname{count})&=\set{\tup{\cname{84},\,\cname{images/id/84/party2.jpg},\,\cname{4}}}\\
		m(\pname{out})&=\set{\tup{\set{}}}\\
		\end{array}$
		\nodepart{two}
		$\I_1=(\mdb^1,\odb^1)$};
		
	\node[state, very thick,label=below:$\snapshot_2$,right of = s1, xshift=10cm] (s2) {%
		$\begin{array}{rl}
		m(\pname{Segments})&=\set{\tup{\cname{84},\,\cname{(204,80)..(270,130)}},\ldots}\\
		m(\pname{count})&=\set{\tup{\cname{84},\,\cname{images/id/84/party2.jpg},\,\cname{3}}}\\
		m(\pname{ready})&=\begin{array}{@{}c@{}}
											\{\langle\cname{84f1},\,\cname{images/id/84/party2\_face1.jpg},\\
											\;\,\cname{84},\,\cname{images/id/84/party2.jpg},\,\cname{3},\,\cname{\set{~}}\rangle\}
											\end{array}\\
		m(\pname{out})&=\set{\tup{\set{}}}\\
		\end{array}$
		\nodepart{two}
		$\I_2=(\mdb^2,\odb^2)$};

		\node[state, very thick,label=below:$\snapshot_2^\prime$,below of = s2, yshift=-3.3cm] (s21) {%
		$\begin{array}{rl}
		m(\pname{Segments})&=\set{\tup{\cname{84},\,\cname{(204,80)..(270,130)}},\ldots}\\
		m(\pname{count})&=\set{\tup{\cname{84},\,\cname{images/id/84/party2.jpg},\,\cname{3}}}\\
		m(\pname{ready})&=\begin{array}{@{}c@{}}
											\{\langle\cname{a12},\,\cname{images/id/84/party2\_face100.jpg},\\
											\;\,\cname{84},\,\cname{images/id/84/party2.jpg},\,\cname{3},\,\cname{\set{~}}\rangle\}
											\end{array}\\
		m(\pname{out})&=\set{\tup{\set{}}}\\
		\end{array}$
		\nodepart{two}
		$\I_2^\prime=(\mdb^{2^\prime},\odb^{2^\prime})$};
		
	\node[state, very thick,label=below:$\snapshot_3$,below of = s21, yshift=-3.3cm] (s3) {%
		$\begin{array}{rl}
		m(\pname{Segments})&=\set{\tup{\cname{84},\,\cname{(50,80)..(125,130)}},\ldots}\\
		m(\pname{count})&=\set{\tup{\cname{84},\,\cname{images/id/84/party2.jpg},\,\cname{3}}}\\
		m(\pname{ready})&=\begin{array}{@{}c@{}}
											\{\langle\cname{84f1},\,\cname{images/id/84/party2\_face1.jpg},\\
											\;\,\cname{84},\,\cname{images/id/84/party2.jpg},\,\cname{3},\,\cname{\set{~}}\rangle\}
											\end{array}\\
		m(\pname{out})&=\set{\tup{\set{}}}\\
		\end{array}$
		\nodepart{two}
		$\I_3=(\mdb^3,\odb^3)$};
		
		\node[draw = none, right of = s2,xshift=6cm] (d1) { \tiny $\bullet\bullet\bullet$};
		\node[draw = none, right of = s21,xshift=6cm] (d2) { \tiny $\bullet\bullet\bullet$};
		\node[draw = none, right of = s3,xshift=6cm] (d3) { \tiny $\bullet\bullet\bullet$};
		\node[draw = none, below of = s3,yshift=1cm,xshift=-8cm] (d4) { \tiny $\bullet\bullet\bullet$};

		\path[very thick] 
		(s1) edge node[above,sloped] {\tname{Split}} (s2)
		(s1) edge node[above,sloped] {\tname{Split}} (s21.west)
		(s1) edge node[above,sloped] {\tname{Split}} (s3.west)
		(s2.east) edge (d1)
		(s2.east) edge (d2)
		(s2.east) edge (d3)
		(s21.east) edge (d1)
		(s21.east) edge (d2)
		(s21.east) edge (d3)
		(s3.east) edge (d1)
		(s3.east) edge (d2)
		(s3.east) edge (d3)
		(s1) edge (d4)
		;
\end{tikzpicture}}
\caption{A transition system of the net from \cref{fig:mmnet-firing}} \label{fig:mmnet-ts}
\end{figure*}
\begin{example}
As stated above, the execution of an \mmnet accounts for the simultaneous progression of its control and data components. Let us consider \mmnet $\net$ from \cref{fig:mmnet-firing}. A transition system capturing its execution semantics starting from snapshot $\snapshot_1$ is shown in \cref{fig:mmnet-ts}. Instances $\mdb^1$ and $\odb^1$ in the multimedia storage instance of $\snapshot_1$ correspond to those from \cref{ex:net-firing}. We also explicitly demonstrate a few possible executions of the net from $\snapshot_1$ under the firing of transition\tname{Split}. Whereas the progression from $\snapshot_1$ to $\snapshot_2$ corresponds to the transition firing discussed in \cref{ex:net-firing}, one may generate infinitely many snapshots similar to $\snapshot_2$, in which the only difference can be noticed in the identifier and address of the extracted sub-image that, according to the semantics of $\nu$-variables, have to contain data that are``locally fresh'' (see snapshot $\snapshot_2^\prime$). Snapshot $\snapshot_3$ in \cref{fig:mmnet-ts} is generated analogously to $\snapshot_2$ by extracting a sub-image with coordinates $\cname{(204,80)..(270,130)}$. 

\end{example}

\subsection{Connection to \dbnets}
It is easy to see that
\mmnets are Turing complete. 
However, this formalism can still be potentially used for checking formal properties of multimedia integration patterns and their compositions. As this paper primarily focuses on developing a modeling formalism, we leave a more in-depth discussion of the formal analysis to the future work and show a key connection between \mmnets and their predecessor \dbnets that, as it has been shown in \cite{MR17} and \cite{RITTER2019101439}, can be used for model-based testing via simulation as well as verification of formal properties such as reachability of a nonempty place.

Formalisms of \dbnets and \mmnets are conceptually quite similar.
The main difference lies in the type of persistent data these formalisms manipulate and the data types they use (as we have stated before, \dbnet data types do not support functions). \dbnets allow for checking the reachability of a nonempty place under restrictions limiting the data types that one can use (namely, only strings and reals) and  the ``size'' of information that can be simultaneously present in the net marking and database instance \cite{MR17}. The same result could be reconstructed for \mmnets with strings and reals (proviso that one uses their definitions from \cite{MR17}) as well as boundedness restrictions imposed over the places and multimedia storage boundedness, and by encoding them into \dbnets. The RDF storage used in \mmnets can be suitably represented in a relational database (\eg \cite{BDKSDUB13} for more details), whereas the SPARQL queries can be translated to SQL as it is suggested in \cite{KRRXZ14}. 
For validating \mmnets, we can leverage the same modular approach used for validating \dbnets in \cite{RITTER2019101439, DBLP:conf/edoc/0001RMRS18}. 
More specifically, one can use CPN Tools (\url{http://cpntools.org/}) for representing the control layer of \mmnets together with queries assigned to view places and actions appearing as code segments attached to transitions, and use its Access/CPN framework for defining extensions that would allow to implement the data manipulation logic running on common RDF / SPARQL frameworks like Apache Jena (\url{jena.apache.org}) and image processing capabilities like OpenCV (\url{opencv.org}), as used in \cite{RitterR17}.

\section{Multimedia Pattern Realization}
\label{sec:realization-analysis}
%
In this section we formalize multimedia integration patterns used in \cref{fig:example} as \mmnets, and thus demonstrate how to model multimedia Message Filter and Content Enricher. The realization of the Splitter can be already found in \cref{sec:mmnets}. We also provide a realization of another important multimedia integration pattern, a Feature Detector, that is, however, not mandatory for our scenario.
Some of the patterns are structurally similar to previous works on textual integration patterns (\eg \cite{DBLP:conf/caise/FahlandG13,RITTER2019101439}).

\subsection{Multimedia Operations}
\label{sec:mm-operations}
We start by outlining various types of multimedia operations used in this section. For simplicity, we consider only the $\typename{jpg}$ type and assume that all the functions that will be formally defined further extend $\sigf_{\typename{jpg}}$. 
Function $\fname{countIMGs}:\mathbb{IMG}\times\mathbb{S}\rightarrow \mathbb{N}$ takes an image together with a feature pattern and counts all sub-images in the given image that correspond to this pattern. If no sub-images have been detected, the function returns $0$.
To detect segments in images, we introduce a function $\fname{detectIMG}:\mathbb{IMG}\times\mathbb{S}\rightarrow \powerset(\mathbb{D})$ that, given an image and a feature, returns a set of segments (of type $\typename{Set_{rect}}$) corresponding to image parts with the detected feature. 
In certain scenarios, it is important to highlight detected objects with a text and/or geometrical shapes. Function $\fname{markIMG}:\mathbb{IMG}\times\mathbb{D}\times\mathbb{S}\times\mathbb{S}\rightarrow\mathbb{IMG}$ is the function that, given an image, segment coordinates, geometrical shape and  color, draws the colored shape around the specified segment in the image. 

%
%

\subsection{Message Filter}
A multimedia Message Filter is supposed to check on incoming messages (carrying both multimedia objects and their metadata), filtering out those that do not match a certain criterion and routing the others to the output channel. 
%
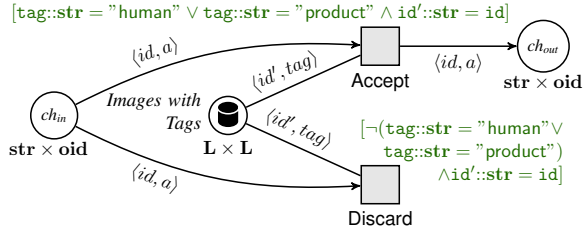
\begin{figure}[h!]
\centering  

\resizebox{.9\hsize}{!}{
\begin{tikzpicture}[->,>=stealth',auto,x=2cm,y=.85cm,thick]

  \node[place,label={[yshift=1mm,xshift=-1mm]below:$\typename{str}\times\oid$}] (in) at (-1.5,-1.5) {$\pname{ch}_\pname{in}$};
 
 	\viewplace{images}{(.1,-1.5)}{left,yshift=.05cm:$\begin{array}{@{}r@{}} \pname{Images with}\\ \pname{Tags}\end{array}$};
 	  \node[draw=none,below of = images,xshift=0cm,yshift=4mm] (type) {$\typename{\LITs}\times\typename{\LITs}$};

	\node[transition, label=below:\tname{Accept}] (yes) at (1.5,0) {};
	\node[draw=none,left of = yes,xshift=-12mm,yshift=6mm] (guard1) {$\gname{\cast{tag}{\typename{str}}=\cname{"human"}\lor \cast{tag}{\typename{str}}=\cname{"product"}\land\cast{id'}{\typename{str}}=id}$};
	
	\node[transition, label=below:\tname{Discard}] (no) at (1.5,-3) {};
	\node[draw=none,above of = no,xshift=1.5cm,yshift=-4mm] (guard2) {$\begin{array}{@{}r@{}}\gbody{[\neg(\cast{tag}{\typename{str}}=\cname{"human"}\lor}\\  \gbody{\cast{tag}{\typename{str}}=\cname{"product"})}\\ \gbody{\land\cast{id'}{\typename{str}}=id]}\end{array}$};

	\node[place,label={[yshift=1mm]below:$\typename{str}\times\typename{oid}$}] (out) at (3,0) {$\pname{ch}_\pname{out}$};

\path
(in) edge[bend left=15] node[above,sloped,pos=.3] {$\tup{id,a}$}  (yes)
(in) edge[bend right=15] node[below,sloped,pos=.3] {$\tup{id,a}$}  (no)
(yes) edge node[below] {$\tup{id,a}$}  (out)
;

\path[-]
(images) edge node[above,pos=0.4,sloped] {$\tup{id',tag}$} (yes)
(images) edge node[above,pos=0.4,sloped] {$\tup{id',tag}$} (no)
;

\end{tikzpicture}
}
\caption{\mmnet control layer of a Message Filter} \label{fig:mmnet_message_filter}
\end{figure}

In \mmnets, this pattern is realized by encoding the filtering condition directly into the net using view places and SPARQL queries attached to them. 
Notice that this realization implicitly requires that the metadata already contain semantic information needed for checking the criterion (\eg triples specifying how many humans/products are present on the pictures).
The net in \cref{fig:mmnet_message_filter} starts by consuming a message from input channel place $\pname{ch}_\pname{in}$ that carries an image identifier and its address in the object storage. 
View place \pname{Images with Tags} is equipped with a query that extracts image identifiers together with tags of objects that they contain. The query itself is defined as $(\tup{?id,?tag},\pattern_1)$, where $\pattern_1$ is as follows:
\begin{align*}
SELECT~ ?id,?tag~ \WHERE \{?id~ \mmdb{containsObj}~?tag\}
\end{align*}
The tags of interest are specified in transition guards that 
realize filtering conditions. If the image with identifier $id$ satisfies the condition, \ie the view place contains a pair in which the first element matches the value carried by $id$ and the second element is either $\cname{"human"}$ or $\cname{"product"}$, then the token with the input message is routed to the output channel by firing transition $\tname{Accept}$. Otherwise, the message gets discarded with transition $\tname{Discard}$.


\subsection{Content Enricher}

\begin{figure}[h!]
\centering 
\vspace{-.8em}
\resizebox{.90\hsize}{!}{
\begin{tikzpicture}[->,>=stealth',auto,x=2.2cm,y=1.2cm,thick]

  \node[place,label={[yshift=1mm,xshift=-1mm]below:$\typename{str}\times\typename{str}\times\oid$}] (in) at (-.5,0) {$\pname{ch}_\pname{in}$};
	
	\node[transition, label=above:\tname{T}] (t1) at (.5,0) {};
	
		\node[place,label={[yshift=1mm]below:$\typename{str}\times\typename{oid}$}] (p1) at (1.8,0) {};
		
		\node[transition, label=above:\tname{Enrich}] (enrich) at (3.4,0) {$\begin{array}{@{}c@{}} \actname{updImage}\\(a,seg)\end{array}$};
		
		\node[place,label={[yshift=0mm]left:$\typename{str}\times\typename{str}$}] (p2) at (.5,-1.7) {};		

		\node[transition, label={[yshift=-1mm]above:\tname{Get Segment}}] (t2) at (1.8,-1.7) {};
		\node[draw=none,left of = t2,xshift=-13mm,yshift=-6mm] (guard1) {$\gbody{[\cast{k'}{\typename{str}}=k\land}\gbody{\cast{id'}{\typename{str}}=id]}$};
		
		 \viewplace{segments}{(1.8,-3)}{left,yshift=-.1cm:\pname{SegmentsByKey}};
  \node[draw=none,below of = segments,xshift=0cm,yshift=4mm] (type) {$\typename{str\times str\times rect}$};

		\node[place,label={[yshift=0mm]right:$\typename{rect}\times\typename{str}$}] (p3) at (3.4,-1.7) {};		

	\node[place,label={[yshift=1mm]below:$\typename{str}\times\typename{oid}$}] (out) at (4.5,0) {$\pname{ch}_\pname{out}$};

\path
(in) edge node[above,pos=.5] {$\tup{k,id,a}$}  (t1)
(t1) edge node[above,pos=.5] {$\tup{id,a}$}  (p1)
(t1) edge node[left,pos=.7] {$\tup{k,id}$}  (p2)
(p2) edge node[above,pos=.3] {$\tup{k,id}$}  (t2)
(t2) edge node[above,pos=.6] {$\tup{\cast{seg}{\typename{rect}},id}$}  (p3)
(p1) edge node[above,pos=.5] {$\tup{id,a}$}  (enrich)
(p3) edge node[right,pos=.3] {$\tup{seg,id}$}  (enrich)
(enrich) edge node[above,pos=.5] {$\tup{id,a}$}  (out)
;

\path[-]
(segments) edge node[right,pos=0.4] {$\tup{id',k',seg}$} (t2)
;

\end{tikzpicture}
}
\caption{\mmnet control layer of a Content Enricher} \label{fig:mmnet_enricher}
\end{figure}
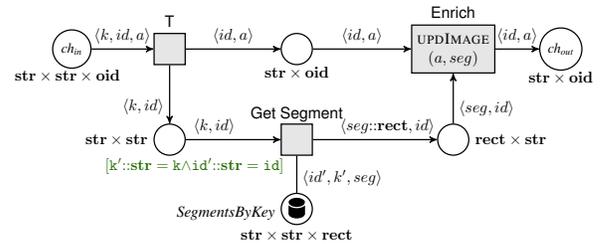
A Content Enricher enriches the content of incoming messages using external sources.
In our case, it utilizes an image identifier to access its metadata and extract information about detected features, using which it then updates the image in the object storage by adding to it extra visual components, as shown in \cref{fig:mmnet_enricher}.

The net starts with a message that contains object identifier $id$, its address $a$ and some feature key $k$ that will be used for acquiring data for the enrichment. Then the net proceeds by splitting the message into two parts and using its part with $id$ and $k$ to get information about a segment that is characterized by $k$ in the metadata storage. To this end, we use view place \pname{SegmentsByKey} that has a query with the following graph pattern attached to it: 
\begin{align*}
\SELECT~ ^*~ \WHERE \{?id~ \mmdb{faceSegment}~?s.\\ ?id~ \mmdb{prodSegment}~?s.\}
\end{align*}
This pattern returns all triples that have $\mmdb{faceSegment}$ and $\mmdb{prodSegment}$ as predicates. The net then allows to choose any segment $seg$ that matches the image identifier and the feature key. Finally, the enrichment step happens when transition \tname{Enrich} gets fired and calls an action assigned to it. This action, called $\actname{updImage}$, updates a (physical) image in the object storage by adding a red oval around some area in it defined by selected segment $seg$. Formally, $\aadd{\actname{updImage}}=(\emptyset,\set{a\callf{\fname{markIMG}(\source(a),seg,\cname{"oval"},\cname{"red"})}})$.


\subsection{Feature Detector}

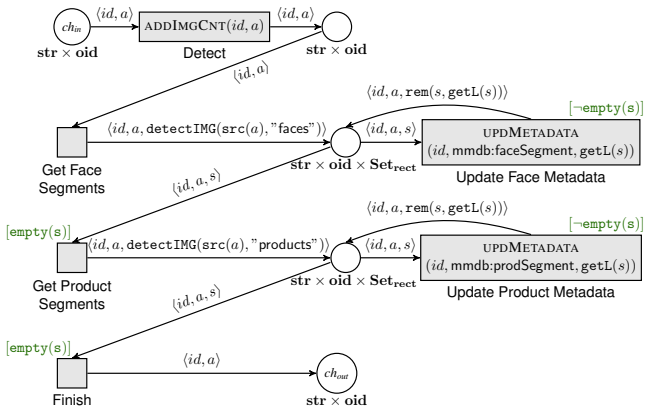
\begin{figure}[h!]
\centering
\resizebox{.99\hsize}{!}{
\begin{tikzpicture}[->,>=stealth',auto,x=2.1cm,y=1.1cm,thick]

  \node[place,label={[yshift=1mm,xshift=-1mm]below:$\typename{str}\times\oid$}] (in) at (0,0) {$\pname{ch}_\pname{in}$};
	\node[transition, label=below:\tname{Detect}] (detect) at (1.5,0) {$\actname{addImgCnt}(id,a)$};
	
	\node[place,label={[yshift=1mm]below:$\typename{str}\times\typename{oid}$}] (detected) at (3,0) {};

	\node[transition, label=below:$\begin{array}{@{}c@{}} \text{\tname{Get Face}}\\ \text{\tname{Segments}}\end{array}$] (getfaces) at (0,-2.5) {};
	
	\node[place,label={[yshift=1mm,xshift=2mm]below:$\typename{str}\times\typename{oid}
\times\typename{Set_{rect}}$}] (faces) at (3.1,-2.5) {};

	\node[transition, label=below:\tname{Update Face Metadata}] (updfaces) at (5.2,-2.5) {$\begin{array}{@{}c@{}} \actname{updMetadata}\\ (id,\mmdb{faceSegment},\fname{getL}(s))\end{array}$ };
	\node[draw=none,right of = updfaces,xshift=8mm,yshift=8mm] (guard1) {$\gname{\neg empty(s)}$};

\node[transition, label=below:$\begin{array}{@{}c@{}} \text{\tname{Get Product}}\\ \text{\tname{Segments}}\end{array}$] (getproducts) at (0,-5) {};
	\node[draw=none,left of = getproducts,xshift=2mm,yshift=6mm] (guard2) {$\gname{empty(s)}$};
	
	\node[place,label={[yshift=1mm,xshift=2mm]below:$\typename{str}\times\typename{oid}
\times\typename{Set_{rect}}$}] (products) at (3.1,-5) {};

\node[transition, label=below:\tname{Update Product Metadata}] (updproducts) at (5.2,-5) {$\begin{array}{@{}c@{}} \actname{updMetadata}\\ (id,\mmdb{prodSegment},\fname{getL}(s))\end{array}$ };
	\node[draw=none,right of = updproducts,xshift=8mm,yshift=8mm] (guard3) {$\gname{\neg empty(s)}$};

	\node[transition, label=below:\tname{Finish}] (finish) at (0,-7.5) {};
	\node[draw=none,left of = finish,xshift=2mm,yshift=6mm] (guard2) {$\gname{empty(s)}$};
	\node[place,label={[yshift=1mm]below:$\typename{str}\times\typename{oid}$}] (out) at (3,-7.5) {$\pname{ch}_\pname{out}$};

\path
(in) edge node[above,pos=.5] {$\tup{id,a}$}  (detect)
(detect) edge node[above,pos=.5] {$\tup{id,a}$}  (detected)
($(detected.west)-(0,.1)$) edge node[below,pos=.3,sloped] {$\tup{id,a}$} (getfaces.north)
(getfaces) edge node[above,pos=.5,xshift=2mm] {$\tup{id,a,\fname{detectIMG}(\source(a),\cname{"faces"})}$} (faces)
(faces) edge node[above,pos=.5] {$\tup{id,a,s}$} (updfaces)
(updfaces.north) edge[bend right = 19] node[above,pos=.5] {$\tup{id,a,\fname{rem}(s,\fname{getL}(s))}$} (faces.north)
($(faces.west)-(0,.1)$) edge node[above,pos=.5,sloped] {$\tup{id,a,s}$} (getproducts.north)
(getproducts) edge node[above,pos=.5,xshift=0mm] {$\tup{id,a,\fname{detectIMG}(\source(a),\cname{"products"})}$} (products)
(products) edge node[above,pos=.5] {$\tup{id,a,s}$} (updproducts)
(updproducts.north) edge[bend right = 19] node[above,pos=.5] {$\tup{id,a,\fname{rem}(s,\fname{getL}(s))}$} (products.north)
($(products.west)-(0,.1)$) edge node[above,pos=.5,sloped] {$\tup{id,a,s}$} (finish.north)
(finish) edge node[above,pos=.5] {$\tup{id,a}$} (out)
;

\path[-]

;

\end{tikzpicture}
}
\caption{The control layer of a \mmnet representing a Feature Detector} \label{fig:mmnet_feature-detector}
\end{figure}
A Feature Detector is a pattern that updates metadata of an object. More specifically, it uses concrete feature classifiers based on which it retrieves data that are later on added to the metadata storage. In our case, this pattern uses a picture identifier to access its metadata and to add information on how many humans and products are in the picture, and, if any have been detected, provides data on the coordinates of segments where human faces as well as products can be found. 

The net starts by executing transition \tname{Detect} that, in turn, calls action $\actname{addImgCnt}$ that has two formal parameters $id$ and $a$, and that upon firing also adds two RDF triples to the metadata storage: $(id,\mmdb{faceCount},\fname{countIMGs}(\source(a),\cast{\cname{"human\,face"}}{\LITs}))$ and $(id,\mmdb{prodCount},\fname{countIMGs}(\source(a),\cast{\cname{"product"}}{\LITs}))$. 

The net then proceeds with updating the metadata storage with the information about coordinates of sub-images that either contain human faces or products. By firing transition \tname{Get Face Segments}, the net generates a set of segments with faces. Until this set is not empty, each of its elements gets removed (using function $\fname{rem}$) and added to the metadata storage with transition \tname{Update Face Metadata}. This transition calls action $\actname{updMetadata}$ that has three formal parameters $\apar{\actname{updMetadata}}=\tup{id,l,seg}$, where $id$ is an image identifier, $l$ is an IRI and $seg$ is a segment. This action does not remove anything and adds to the metadata storage only one triple $(\cast{id}{\LITs},\cast{l}{\IRIs},\cast{seg}{\LITs})$. In case of \tname{Update Face Metadata}, $\actname{updMetadata}$  adds to the metadata (of an image with identifier $id$) information about one face segment taken from set $s$ that is specified with IRI $\mmdb{faceSegment}$.

When the set of segments is empty, the net performs the similar procedure with product segments. That is, it first gets a set of all the product segments by firing \tname{Get Product Segments}, and then updates the metadata storage by consecutively firing \tname{Update Product Metadata} for each segment from the set. After all the updates are done, the net finishes its computation by firing \tname{Finish} and placing a token with the image identifier and address into place $\pname{ch}_\pname{out}$.

\subsection{Discussion}

We have demonstrated how some of the most important (multimedia) integration patterns can be formally represented in our formalism.
 Notice that, similarly to \cite{DBLP:conf/caise/FahlandG13} and \cite{RITTER2019101439,DBLP:conf/edoc/0001RMRS18}, every pattern should be equipped with input and output channels. This is needed to ensure their flawless, message-based composition: in case of two connected pattern formalizations, the output channel of the first one should have the same type as the input channel of the second. 
In the scenario described in \cref{ex:example} and depicted in \cref{fig:example}, the whole process essentially represents a sequential composition of integration patterns and thus can be seamlessly implemented by following the order of patterns in \cref{fig:example} and by ``fusing'' output and input channels of two neighboring \mmnet pattern representations.  Indeed, by mapping every task into its corresponding Petri net-based representation, one can easily build a model formalizing the entire SAP Social Intelligence scenario. 
Notice, however, that, strictly speaking, textual patterns in \cref{fig:example} would need to be formalized by either using CPNs~\cite{DBLP:conf/caise/FahlandG13} or \dbnets~\cite{RITTER2019101439}. In the first case, the control layer of \mmnets captures the whole class of CPNs, resulting in their seamless adoption when implementing multimedia EAI scenarios. In case of \dbnets, however, one would need to study in more detail how to implement the multi-model integration scenarios~\cite{RitterR17} that use both relational and multimedia databases. In our case, the textual Splitter and Content Enricher patterns do not require any database access and thus the whole scenario can be implemented using \mmnets.


Using the scenario in \cref{ex:example}, we also identified a suitable way to formally represent multimedia manipulating functions and semantic operations using data types (cf. \cref{sec:mm-operations}), SPARQL queries (cf. \cref{sec:mm-storage}) and actions/queries in the data logic layer (cf. \cref{sec:data-logic}). Derived functions/queries/actions are providing the full coverage of physical and logical operations in \cref{tab:imageOperations} for studied patterns, and, moreover, can be seen as pattern-agnostic since they may be re-used in other image-based scenarios (formalized using \mmnets) as construction primitives, akin to pattern implementations. 

\section{Related Work}
\label{sec:relatedwork}
Ritter et al. \cite{Ritter201736} showed that, for structured data, the only existing formalization of integration patterns was studied by Fahland et al. \cite{DBLP:conf/caise/FahlandG13} using CPNs, and that was further extended to cover a wider range of integration patterns with more refined requirements in \cite{RITTER2019101439} using (timed) \dbnets.
However, when considering multimedia integration patterns (cf. \cite{RitterR17}), as briefly introduced in \cref{sec:analysis}, these works cannot be used directly (cf. requirements in \cref{tab:requirements}), due to their lack of multimedia data operations (cf. REQ1(a)), semantic operations (cf. REQ-1(b)), and partially their storage (cf. REQ-1(c)).
To the best of our knowledge, there have been no attempts to formalize multimedia integration patterns (cf. \cite{Ritter201736}). 

\labeltitle{Formalisms for integration patterns} Although the scenario in \cref{ex:example} is captured in BPMN~\cite{ritter2016exception}, this modeling language is not suitable for our requirements (especially REQs-1(a--c), 2).
Yet, BPMN diagrams can be formally represented using Petri nets \cite{dijkman2007formal}. 
Petri nets offer a good trade-off between user-friendly graphical modeling and a toolbox for formal analysis of produced models. There are many data-aware extensions of Petri nets (\eg \cite{PWOB19,MR17,de2017add,BadouelHM16}) allowing to account for more complex, structured data. However, they cannot be readily used for representing multimedia EAI for reasons similar to those discussed above. 
Alternatively, it is possible to study other modeling requirements in which the multimedia message is treated as the first-class citizen, while the object storage is simply left out.
Under this assumption, one could use the formalism of Petri nets with structured data~\cite{BadouelHM16} (StDNs for short) for modeling multimedia EAI as tokens in it carry XML documents that, in turn, could be used for representing multimedia metadata -- the core concept of the multimedia message. The authors also delineate restrictions required for the decidability of such properties as termination, coverability and boundedness. However, the formalism still does not account for persistent data (violation of REQ-1(d) that is crucial for some patterns) and would need to be extended with the support of functions.


Mederly et al. \cite{mederly2009construction} studied an approach for formalizing integration patterns, in which messages are first-order formulas and patterns are operations that add and delete messages, and that uses AI planning for finding an integration process with a minimal number of components.
While this approach shares the formalization objective, \mmnets apply to a broader set of objectives (\eg formal analysis, simulation) and cover multimedia data, semantics and storage (cf. REQs-1(a--c)). 

\labeltitle{Multimedia data} 
The approach for storing and querying multimedia data employed in this work is similar to the one in the OCAPI system \cite{clement1990integration}, which was developed for the semantic integration of image data using knowledge bases. 
%
%
%
%
%
%
%
Retrieval of multimedia information from (distributed) databases is covered in \cite{DBLP:journals/tkde/ChangDPV07}.
The multimedia semantics are represented by semantic attributes based on extended generalized icons with a logical and physical representation on a database.
While our approach separates these different representations as well, \cite{DBLP:journals/tkde/ChangDPV07} targets extended normal forms and functional dependencies between different attributes and does not define user interaction with the multimedia semantics on a business application-relevant feature level that could be used for message processing.
More recently, \cite{DBLP:conf/vldb/OzsuLON01} developed an image similarity query mechanism in the area of multimedia queries in multimedia databases.
While no query syntax is provided, the approach could be used to formulate decisions based on image similarity.
\section{Conclusions}
\label{sec:conclusions}
The previous work on EAI with multimedia data \cite{RitterR17,Ritter201736} pinpointed two main issues in the domain. On the one hand, it argued that the integration patterns for multimedia scenarios are still not fully investigated, which  can cause their wrong adoption in the EAI development stack. On the other hand, there is no formalization of these patterns that would allow to minimize design-time mistakes, facilitate the model-driven development and provide possibility for checking correctness of the implemented multimedia EAI scenarios. In this work we focused on the second issue and distilled a list of requirements (cf. \cref{tab:requirements}) for the formal representation of multimedia EAI. To address these requirements, we studied a formalism of \mmnets that marries CPNs and multimedia databases, and that allows to specify operations that manipulate both the multimedia objects and their metadata.
The paper also presents how \mmnets can be used for formalizing some of the most frequently used multimedia integration patterns. 
We believe that the formalism studied in this paper can also provide more insights on engineering multimedia EAI. 
Currently we are working on developing a CPN Tools-based prototype for modeling and simulating \mmnets and studying more in-depth formal analysis of \mmnet models. In the future, it would be also interesting to study a domain-independent language for representing multimedia manipulation functions and creating their repository in order to facilitate their adoption in different multimedia EAI scenarios.


%
%

\bibliographystyle{abbrv}
\bibliography{mybib}

\end{document}